# ESTIMATION AND NAVIGATION METHODS WITH LIMITED INFORMATION FOR AUTONOMOUS URBAN DRIVING

A Dissertation

Presented to the Faculty of the Graduate School

of Cornell University

In Partial Fulfillment of the Requirements for the Degree of

Doctor of Philosophy

by

Jordan Bradford Chipka

December 2018

ESTIMATION AND NAVIGATION METHODS WITH
LIMITED INFORMATION FOR AUTONOMOUS URBAN DRIVING

Jordan Bradford Chipka, Ph. D.

Cornell University 2018

Urban environments offer a challenging scenario for autonomous driving. Globally localizing information, such as a GPS signal, can be unreliable due to signal shadowing and multipath errors. Detailed *a priori* maps of the environment with sufficient information for autonomous navigation typically require driving the area multiple times to collect large amounts of data, substantial post-processing on that data to obtain the map, and then maintaining updates on the map as the environment changes. This dissertation addresses the issue of autonomous driving in an urban environment by investigating algorithms and an architecture to enable fully functional autonomous driving with limited information.

In Chapter 2, an algorithm to autonomously navigate urban roadways with little to no reliance on an a priori map or GPS is developed. Localization is performed with an extended Kalman filter with odometry, compass, and sparse landmark measurement updates. Navigation is accomplished by a compass-based navigation control law. Key results from Monte Carlo studies show success rates of urban navigation under different environmental conditions. Experiments validate the simulated results and demonstrate that, for given test conditions, an expected range can be found for a given success rate.

Chapter 3 develops an approach to detecting and estimating key roadway features, which are then used to create an understanding of the static scene around the vehicle. The primary focus of this study is specifically on intersections, given their complexity compared to other scenes and their importance to navigation. Using a test vehicle equipped with a vision system, odometry and vision data is collected for a variety of intersections under diverse conditions. Experimental results are then obtained using computer vision and estimation techniques. These results demonstrate the ability to probabilistically infer key features of an intersection as the vehicle approaches the intersection, in real-time.

In separate earlier research, a novel, meso-scale hydraulic actuator characterization test platform, termed a Linear Hydraulic Actuator Characterization Device (LHACD), is developed. This work is detailed in Chapter 1. The LHACD is applied to testing McKibben artificial muscles and is used to show the energy savings due to the implementation of a variable recruitment muscle control scheme. The LHACD is a hydraulic linear dynamometer that offers the ability to experimentally validate the muscles' performance and energetic characteristics. For instance, the McKibben muscles' quasi-static force-stroke capabilities, as well as the power savings of a variable recruitment control scheme, are measured and presented in this work. Moreover, the development and fabrication of this highly versatile characterization test platform for hydraulic actuators is described in this chapter, and characterization test results and efficiency study results are presented.

# BIOGRAPHICAL SKETCH

Jordan Bradford Chipka attended Rose-Hulman Institute of Technology and graduated in 2013 with a Bachelor of Science degree in Mechanical Engineering. He then began his graduate studies in the Sibley School of Mechanical and Aerospace Engineering at Cornell University in the fall of 2013. He joined the Laboratory of Intelligent Machine Systems (LIMS), advised by Dr. Ephrahim Garcia, and studied efficient actuation methods for walking robots. During this time, Jordan was awarded a National Science Foundation (NSF) Graduate Research Fellowship. After one year of working in LIMS, Jordan changed labs and advisers due to the death of Dr. Garcia. Following a transition period of about one year, Jordan joined the Autonomous Systems Lab (ASL), advised by Dr. Mark Campbell, and studied topics relating to autonomous driving. In the fall of 2016, Jordan completed the requirements to receive his Master of Science degree. In the fall of 2018, Jordan submitted his dissertation for the December 2018 conferral. During his graduate studies, Jordan authored the following papers.

- **J. Chipka** and M. Campbell, "Estimation and navigation methods with limited information for autonomous urban driving," *Journal of Field Robotics*, 2018. *[In Preparation]*
- **J. Chipka** and M. Campbell, "Autonomous urban localization and navigation with limited information," *IEEE Intelligent Vehicles Symposium*, 2018.
- **J. Chipka**, M. Meller, A. Volkov, M. Bryant, and E. Garcia, "Linear dynamometer testing of hydraulic artificial muscles with variable recruitment," *JIMSS*, 2016.
- M. Meller, **J. Chipka**, A. Volkov, M. Bryant, and E. Garcia, "Improving actuation efficiency through variable recruitment hydraulic McKibben muscles: modeling, orderly recruitment control, and experiments," *Bioinspiration & Biomimetics*, 2016.
- E. Ball, M. Meller, **J. Chipka**, and E. Garcia, "Modeling and testing of a knitted-sleeve fluidic artificial muscle," *SMS*, 2016.

# ACKNOWLEDGMENTS

This dissertation has been supported by the NSF GRFP grant DGE1144153 and NSF IIS-1724282.

I am very grateful to many people who have been instrumental to me over the past several years during my PhD studies. Those who deserve special mention are:

- Professor Mark Campbell for his mentorship and guidance.

- Professor Hadas Kress-Gazit and Professor Bart Selman for their teaching and support.

- Professor Ephrahim Garcia for sharing his wisdom early in my PhD studies.

- My lab mates at the Autonomous Systems Lab and the Laboratory of Intelligent Machine Systems, notably Mike Meller, Boris Kogan, Peter Radecki, and Daniel Lee for all the assistance that they gave me throughout my studies.

- My loving wife, Miya, and the rest of my family for their unyielding love and support.

- My pastor at Cornell and his family for helping to build my faith during this long and often difficult endeavor.

TABLE OF CONTENTS

INTRODUCTION

In the past decade, the autonomous driving industry has experienced rapid growth and development of key technologies. However, despite this development, even state-of-the-art technologies still struggle in certain difficult driving scenarios, such as driving in a dense urban environment. Urban environments present a number of unique difficulties in regard to localization, mapping, and planning. For instance, multipath errors and signal shadowing, often experienced in dense urban canyons, make positioning systems based on GPS an unreliable information source. Furthermore, urban areas present difficulties in terms of acquiring and maintaining the highly detailed maps needed for autonomous navigation. These maps, which are expensive to obtain in terms of time, money, and resources to begin with, must undergo continual maintenance to account for the persistent changes associated with rapidly developing urban areas.

For all the above reasons, the work presented in this dissertation investigates innovative estimation and navigation methods for autonomous urban driving, which only require limited *a prior* information. The techniques discussed approach the problem of autonomous urban driving without the use of prior detailed map information or GPS measurements. Not only would a weaker dependence on these data sources mitigate the difficulties discussed above, but it would also help to mitigate other potential problems, such as malicious cybersecurity attacks.

In Chapter 1 of this dissertation, separate earlier research is discussed in which a novel hydraulic actuator characterization device is developed. This device is used to test McKibben artificial muscles in order to experimentally validate the muscles' performance and energetic characteristics, as well as demonstrate the energy savings due the implementation of a variable recruitment control scheme. Details of the development and fabrication of this highly versatile characterization test platform, as well as the characterization test results and efficiency study results, are presented in this chapter.

Chapter 2 of this dissertation presents an algorithm to autonomously navigate urban roadways with little to no dependence on prior detailed map information or GPS measurements. An extended Kalman filter with odometry, compass, and sparse landmark measurement updates is utilized for localization. Navigation is dictated by a compass-based navigation control law. Key results from Monte Carlo studies demonstrate success rates of urban navigation under various environmental conditions. Field tests validate the simulated results and demonstrate that an expected range can be found at a specified success rate for given test conditions.

In the third and final chapter of this dissertation, an approach to detecting and estimating key roadway features to create an understanding of the static scene around the vehicle, is developed. This chapter focuses primarily on intersections due to their complexity compared to other scenes and their importance to navigation. Odometry and vision data is collected for a variety of intersections under diverse conditions using Cornell University's autonomous test vehicle equipped with a vision system. Using computer vision and estimation techniques, experimental results are obtained and used to demonstrate the ability to probabilistically infer key features of an intersection as the vehicle approaches the intersection, in real-time.

CHAPTER 1

LINEAR DYNAMOMETER TESTING OF HYDRAULIC ARTIFICIAL MUSCLES WITH

VARIABLE RECRUITMENT

*Introduction to Hydraulic Actuation*

Hydraulic actuation is a popular choice when a task calls for consistent, large amounts of power output. However, for systems and applications that require short bursts of high power, followed by a period of low power, the inefficiencies of hydraulic power systems become apparent. One system that fits this description is a legged robot. Legged robots have become increasingly popular in today's research community [1], [2], [3]. [4], as designers strive to develop more human-like robots that can negotiate varied terrain or use human tools. However, a major difficulty in the advancement of legged robots and many other meso-scale hydraulic systems is poor actuation efficiency [5]. Thus, this study uses legged robots as the motivating application in our aim to improve actuation efficiency for meso-scale hydraulics.

For hydraulic applications, efficiency is simply the ratio of mechanical output power to fluid input power for a given actuator. Poor actuation efficiency severely restricts the capabilities of a hydraulic system, which in turn limits its practicality. For instance, poor actuation efficiency confines the range of an untethered walking robot, which is one of the major dilemmas in this area of research. This study examines the issue of hydraulic actuation inefficiencies for meso-scale systems through the development of a unique test apparatus, termed a Linear Hydraulic Actuator Characterization Device (LHACD), to measure and validate energy saving techniques for hydraulic actuators. This chapter specifically explores bio-inspired hydraulic systems and the use of McKibben artificial muscles; however, the testing method discussed in this chapter, as well as the energy-saving techniques, can be extended to a wide variety of hydraulic actuators and systems. This chapter extends the initial work described in [6] to include the fully developed hydraulic system, a description of the current state of the controller for the LHACD, characterization testing of the LHACD system and McKibben muscles, and finally application of the LHACD to experimentally quantify the efficiency of a variable recruitment hydraulic artificial muscle system. In addition to the novelty of the LHACD apparatus, this chapter presents the first experimental implementation of online variable recruitment in an artificial muscle system.

*McKibben Artificial Muscles*

The application of interest for this study is the use of McKibben artificial muscles for bipedal robots. Compared to traditional hydraulic cylinders, McKibben muscles, also known as fluidic artificial muscles (FAMs), provide a lightweight and more compliant form of actuation. Furthermore, they have high force capabilities and exhibit force-stroke profiles similar to skeletal muscle [7]. FAMs are comprised of an inner bladder that is held within a helically-woven mesh. The bladder is activated by pressurized fluid that enters from one end of the muscle. The other end of the muscle is plugged, and both ends are rigidly attached to the structure that is being actuated.

McKibben artificial muscles are used both pneumatically and hydraulically in the research community, however, pneumatic actuation is the predominant method used. Both hydraulic and pneumatic artificial muscles have their own distinct benefits. Therefore, these benefits must be weighed in light of the intended application in order to choose the appropriate working fluid for the artificial muscles. Pneumatic actuation offers the convenient ability to vent air to the environment and significantly increases the actuator's compliance [8]. In addition to actuator compliance, pneumatic actuators commonly operate at relatively low pressures. Also, it is, in general, less difficult to maintain a clean working environment with pneumatic actuation, as opposed to hydraulic actuation. For all of these reasons, pneumatic systems are typically considered easier to work with when compared to hydraulic systems. However, hydraulic actuation offers increased force output, efficiency, and improved response time. Therefore, since this study is specifically focused on improving the efficiency of meso-scale actuation, hydraulic artificial muscles were chosen for this research. This decision primarily stems from the relatively-high compressibility of air, which results in lower operating efficiency for valve-controlled pneumatic systems, as well as a slower response time and poor position control authority [9]. In regards to artificial muscles, Meller and Bryant et al. found that, at 75 psi, hydraulic muscles with a latex bladder operate at roughly 60 percent efficiency, while pneumatic muscles with a latex bladder operate at roughly 25 percent [10]. This comes as a result of very little energy being lost while compressing the working fluid, given that hydraulic fluid has a much higher bulk modulus than air. Finally, due to hydraulic fluid's higher bulk modulus, positioning precision and actuator stiffness are also increased [11].

Select research groups have utilized McKibben muscles hydraulically [9], [10], [12], [13], [14]. However, these actuators have not yet been fully characterized in the literature. More specifically, there is limited research to determine the manner in which hydraulic McKibben muscles behave beyond simple quasi-static force-stroke curve measurements. An adequately versatile testing platform on which to characterize these actuators would allow for more exhaustive tests to be performed. This study aims to provide such a device.

*Variable Recruitment Actuation Scheme*

In an effort to reduce the inefficiencies associated with hydraulic actuation for robotics, the issue of valve throttling is considered. Traditionally, the size of a hydraulic actuator is determined by the maximum force output that it needs to provide. Then, through valve throttling, the flow and pressure to the actuator are regulated to meet the force requirement for the specific task. This proves to be vastly inefficient, as valve throttling results in a power loss in the form of expended heat [15]. To combat this difficulty, biology is used as inspiration. In a manner similar to how skeletal muscle fibers are selectively activated to meet the force requirement for a specific task, the hydraulic artificial muscles are bundled so that any number of artificial muscles could be selected to meet the force requirement. This novel actuation strategy was first studied in Bryant et al. [16], and has more recently been pursued by others [17]. This method of actuation, called selective or variable recruitment, requires only a subset of the actuators to be active. For instance, when little mechanical power is required to be delivered by the hydraulic artificial muscles, then only a small number of muscles are activated. With only a small number of muscles active, the muscles operate at a higher pressure for a given force requirement due to the decrease in the total effective cross-sectional area of the muscles (i.e. fewer muscles are active, so less cross-sectional area is used). This case is more efficient than activating more muscles and operating at a lower pressure because less throttling loss is incurred [16]. When more mechanical power is required from the McKibben muscles, more artificial muscles are activated. Thus, this actuation scheme addresses the hydraulic inefficiencies associated with valve throttling by selectively recruiting only the number of artificial muscles required for a specific task. While a variable recruitment hydraulic actuation scheme has the potential to offer large power savings, to date this concept has only been investigated theoretically [16], [18], [19], with quasi-static experiments [16], and with manual, offline recruitment control [17]. The research community is currently lacking in studies that demonstrate on-line variable recruitment of hydraulic actuators and experimentally quantify the power savings that it can provide. Primarily, this is due to the lack of a sufficiently adaptable test bed on which to conduct these experiments and measure the relevant actuator and flow parameters. This study was aimed to develop such a test platform and carry out novel tests to demonstrate the capabilities of hydraulic artificial muscles using a variable recruitment actuation scheme.

*Linear Hydraulic Actuator Characterization Device (LHACD)*

Hydraulic actuator test platforms have been developed both in industry, as well as academia, for actuator characterization and controller development. Some of these test platforms simply use static weight to provide a disturbance to the test actuator [20]. Other test beds use a drive actuator to provide the disturbance. These drive actuators include devices such as a motor and pulley [21], another hydraulic artificial muscle [22], and a traditional hydraulic cylinder [23], [24], [25], [26]. The test platforms using a traditional hydraulic cylinder as the disturbance generator are similar to the LHACD, except that they are designed for a specific set of tests to be performed on a particular test actuator, and so they do not offer the same versatility of the LHACD. Finally, other research groups simply use a universal testing machine to test their actuators [27]. The LHACD consists of a test actuator that is subjected to a disturbance generated by a drive actuator. This drive actuator imparts a dynamic load to the test actuator, while data is being collected from various sensors on the LHACD. Although conventional hydraulic linear dynamometers typically test traditional actuators such as hydraulic cylinders and linear motors, this study extends the use of the LHACD to artificial muscles. Since McKibben muscles are also linear actuators, the LHACD is a useful test bed for muscle characterization and controller development.

In this chapter, we develop the LHACD and utilize it to investigate variable recruitment hydraulic artificial muscles for meso-scale hydraulic systems, such as walking robots. First, the LHACD is used to develop a simple controller that prescribes force and stroke trajectories to the artificial muscles and incorporates the variable recruitment actuation scheme. The LHACD drive cylinder's dynamic characteristics, including frequency response, position response, and open-loop gain are then quantified through several tests. We then proceed to present quasi-static force-stroke characterization tests on the in-house fabricated hydraulic artificial muscles to be used in the variable recruitment bundle. Once these characterization tests are complete, we present efficiency testing of an artificial muscle bundle using a variable recruitment actuation scheme to quantify the power savings for meso-scale hydraulic systems. This is done by experimentally comparing the variable recruitment actuation case to the case of tracking the same force-stroke trajectory with all muscles in the bundle activated.

This chapter is organized in the following manner. First, a discussion of the McKibben artificial muscles used in this study is presented. Also, the custom-fabricated McKibben muscles, the LHACD hydraulic system, as well as the controller implemented on the LHACD are presented. Next, we present the experimental procedure and results from the various tests carried out to characterize the LHACD and the McKibben artificial muscles. Finally, we demonstrate an online variable recruitment actuation scheme in an artificial muscle system to show the power savings that it provides.

***McKibben Artificial Muscles, LHACD, and Hydraulic System Test Setup***

*McKibben Artificial Muscles*

For this study, we developed and tested custom hydraulic artificial muscles. An image portraying the construction of these muscles is seen in Fig. 1 and an image of a completed muscle bundle is shown in Fig. 2. The muscles consist of an inelastic, pleated LDPE bladder, which is sleeved through a Kevlar mesh and then clamped on the ends with hydraulic hose fittings and rubber spacers. The wind angle for the Kevlar mesh used in this study was approximately 29 degrees. Typically, pneumatic artificial muscles use natural rubbers, such as latex, for the bladder [28]. However, most natural rubbers are not compatible with hydraulic fluid. Furthermore, an elastomeric bladder is less energy efficient than an inelastic bladder due to the energy lost from stretching the elastic bladder [10]. Meller et al. found that the use of an LDPE bladder increased efficiency by over 33 percent compared to the use of a latex bladder at 75 psi [10]. Thus, we employed hydraulic McKibben muscles with an inelastic LDPE bladder for this study.

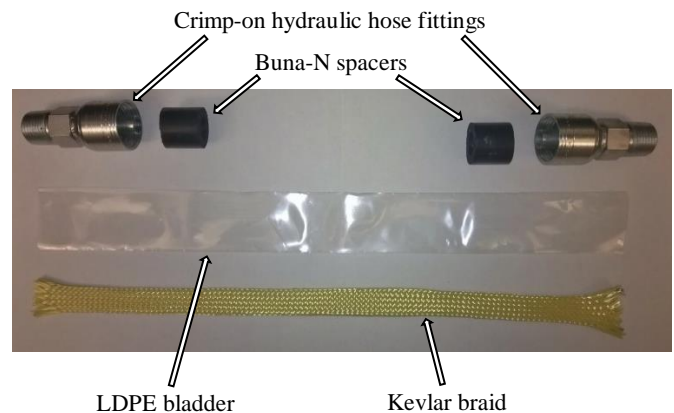

Fig. 1. Materials used for the construction of the custom hydraulic artificial muscles used for this research.

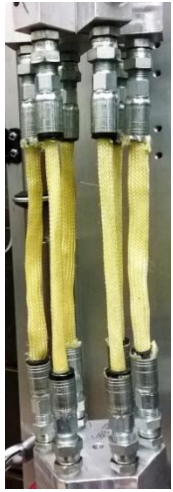

Fig. 2. A completed muscle bundle of six artificial muscles configured in a hexagonal pattern that was used for variable recruitment efficiency testing.

Since the motivating application of our research is meso-scale hydraulic systems, such as walking robots, we used load cell data from tests performed on an in-house designed and constructed walking robot to provide an approximate measure of the force magnitude that the McKibben muscles must generate [6]. Then, we conducted an analysis using the relationship between output force, pressure, and the muscle's geometry, found in equation (1), to determine the appropriate dimensions for our muscles [7]. This analysis sized muscles to replace each double-acting hydraulic cylinder currently on our robot with an antagonistic pair of single-chamber McKibben muscles. For this study, muscles with an initial length of 8 inches and an initial radius of 0.25 inches were used, corresponding to the size required for the robot rear hip actuator.

$$F = (\pi r_0^2)P[a(1-\epsilon)^2 - b] \qquad (1)$$

$$a = \frac{3}{\tan^2(\alpha_0)} \qquad (2)$$

$$b = \frac{1}{\sin^2(\alpha_0)} \tag{3}$$

*Hydraulic System*

The developed LHACD setup allowed for two classes of tests to be performed: 1) single actuator tests, and 2) actuator bundle tests with up to six McKibben muscles in a parallel configuration (as seen in Fig. 2). For the muscle bundle tests, the muscles are fluidically connected in pairs by a custom variable recruitment manifold. Each muscle pair is individually controlled by a servo valve, configured for a single-acting actuator. The LHACD's hydraulic drive cylinder, used to exert external loads or disturbances on the muscles, is controlled by an additional servo valve, configured for a double acting cylinder. The cylinder control valve is supplied by the high pressure line of the LHACD's hydraulic system, set by a relief valve at 600 psi. Generally, hydraulic systems operate at a much higher pressure, typically near 3,000 psi; however, due to our focus on meso-scale, bio-inspired hydraulic systems with artificial muscles, an intermediate operating pressure was chosen. Therefore, a hydraulic power unit with a maximum supply pressure of 600 psi was used. Since the McKibben muscles tested are constructed for lower pressure, a reducing relief valve is used to drop the supply pressure for their control valves to 250 psi. For safety purposes, adjustable emergency relief valves were installed on each muscle control line and both ports of the drive cylinder.

The LHACD system is outfitted with a suite of sensors for both data collection and control purposes. In the mechanical measurement domain, the stroke and tensile load of the artificial muscles are tracked by an LVDT and load cell, respectively, which also serve as the feedback sensors to the controller. In the hydraulic domain, five pressure transducers measure the pressure in each of the muscle pairs, as well as in the drive cylinder's supply and return lines, and a flow meter measures the total volumetric flow rate supplied to the muscles. The pressure sensors and flow meter are used to obtain the necessary measurements to calculate the fluidic power required by the McKibben muscles. Furthermore, the pressure data is used to provide feedback to the controller as well. The LHACD utilizes a Quanser QPIDe data acquisition device as the control board for the system. The QPIDe control board uses the capabilities of Simulink, which allows for future iterations to be easily made to the controller. An annotated view of the entire hydraulic system can be seen in Fig. 3. A list of the equipment used in the hydraulic system for the LHACD is shown in Table 1. Fig. 4 shows a schematic of the simplified hydraulic circuit for the LHACD.

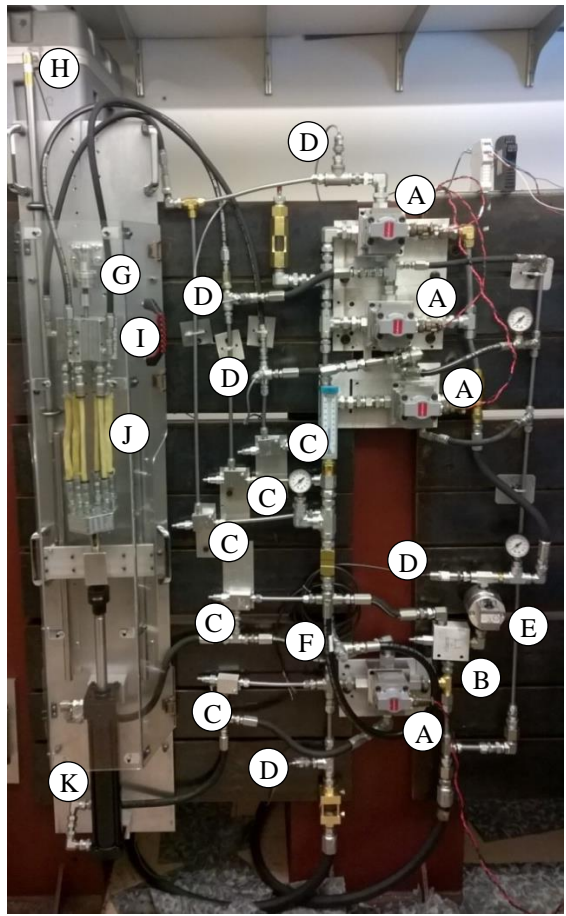

Fig. 3. Annotated picture of the hydraulic system used to test the hydraulic artificial muscles on the LHACD. The letters in the picture correspond to the letters in Table 1.

Table 1. List of equipment for the LHACD

| | |
|---|---|
| A) | Servo valves (x4) (Moog G761) |
| B) | Reducing relief valve (Sun Hydraulics PRDBLEV) |
| C) | Relief valves (x5) (Sun Hydraulics RDBALDV) |
| D) | Pressure transducers (x4) (Omega PX409) and (x1) (Measurement Specialties MSP3000) |
| E) | Flow meter (Omega FPD2000) |
| F) | Temperature sensor (Omega M12TXC) |
| G) | Load cell (Transducer Techniques SSM-1K) |
| H) | LVDT (RDP Group ACT) |
| I) | Variable recruitment manifold |
| J) | McKibben muscle bundle |
| K) | High-pressure drive cylinder (TRD Manufacturing MH, 2.5-inch bore, 1-inch rod diameter) |
| L) | Hydraulic power unit (not pictured) (Concentric GC9500) |
| M) | Control board (not pictured) (Quanser QPIDe) |

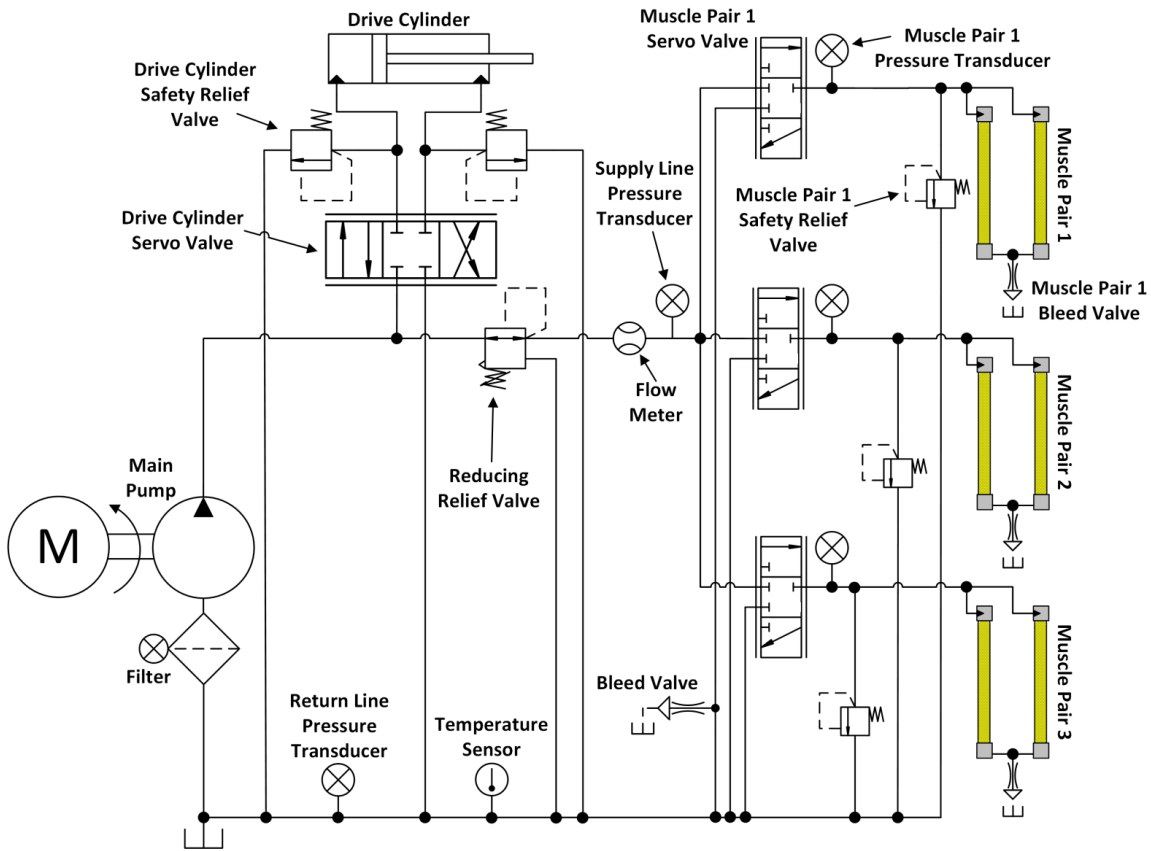

Fig. 4. A schematic of the simplified hydraulic circuit for the LHACD. Pilot lines, safety loopback valve, ball valves, flow sights, and visual gauges are not shown.

*Control System*

Simple hybrid PID control systems were implemented to regulate the flow through the servo valves supplying the drive cylinder and artificial muscles. The discrete aspect of the controllers enables variable recruitment through an output on-off signal to the individual muscle valves to recruit the appropriate number of muscles, as shown in the variable recruitment selector block in Fig. 5. Therefore, the overall control system consists of four main PID control loops, with three of the PID loops used to control each of the artificial muscle pairs, and the fourth loop used to control the drive cylinder. A high-level view of the control system is shown in Fig. 5. As stated in the previous section, the hardware used to control the LHACD is a Quanser QPIDe data acquisition device using the capabilities of Simulink.

The inputs to the PID control loops are the desired position and force profiles for the McKibben muscles. For our current implementation, position is the controlled variable for the drive actuator's PID controller, while force is the controlled variable for the three PID controllers for the artificial muscle bundle. From the feedback data obtained from the sensors, an error in the desired position and force are obtained. These values then feed into their corresponding PID controllers, which then output the appropriate signal related to the error between the prescribed trajectories and measured trajectories. These signals are then sent to the corresponding valve, which opens or closes the orifice the appropriate amount to better match the desired force or position.

While this relatively simple control approach was found to be adequate for the force and stroke trajectories studied in this first demonstration of online variable recruitment, the LHACD provides a test platform on which more advanced controller development can be performed for future studies. This will be needed since simple PID controllers will likely not provide the adequate control authority required for many applications, due to the highly non-linear behavior of the McKibben artificial muscles, and thus improved controllers will be the subject of future investigations.

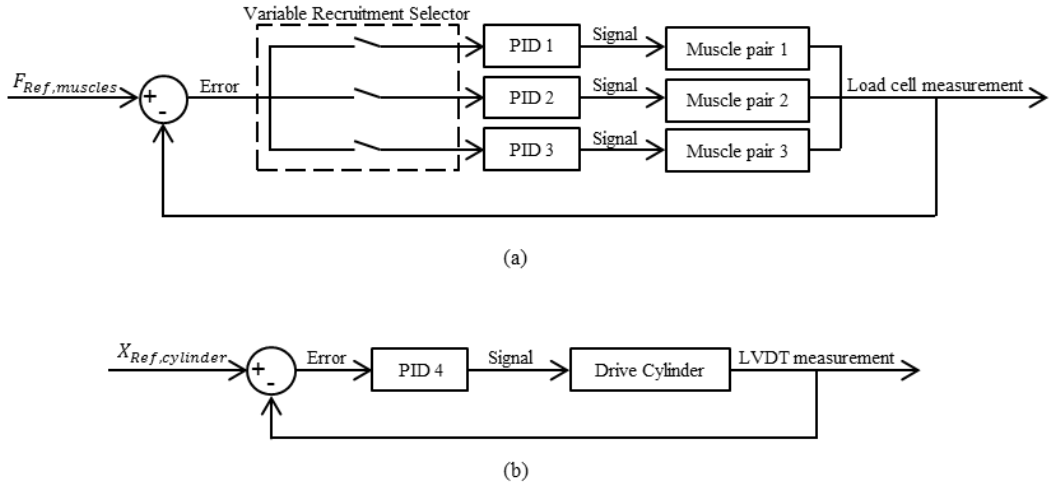

Fig. 5. High-level view of the control system. (a) System of controllers for the artificial muscles. (b) Controller for the drive cylinder.

To incorporate the concept of online variable recruitment, we assumed a simple switching scheme. Three conditional switches were added in software to activate or deactivate each muscle pair based on the instantaneous force requirements. As the required force increases above a recruitment threshold value, a switch closes to activate another pair of artificial muscles. Then, as the required force decreases, the switch opens to deactivate the muscle pair. These discrete switches control whether each PID control loop is active, and therefore, whether each artificial muscle pair is active as well.

One challenging aspect of testing artificial muscles on the LHACD is that the capability of controlling both the load and stroke of the McKibben artificial muscles is needed. Based on the dimensions of the LHACD, and the knowledge that the drive cylinder and artificial muscles are rigidly attached, the stroke of the McKibben muscles can be prescribed by controlling the high pressure drive cylinder's position using the PID controller shown in Fig. 5(b). Then, to control the load in the McKibben muscles, the pressures applied to the McKibben muscles are controlled using the variable recruitment selector and the system of PID controllers shown in Fig. 5(a). This is due to the fact that at a given stroke value, the force generated by a McKibben muscle is directly related to the operating pressure, as seen in equation (1). Alternatively, this control problem can also be approached by controlling the position of the McKibben muscles to achieve the desired stroke profile, while the desired force profile is prescribed to the drive cylinder, which then acts as a disturbance generator. This versatility to prescribe specified force and stroke trajectories to a given test actuator exemplifies the benefits of a test platform such as the LHACD.

*Theoretical Evidence for Variable Recruitment Efficiency Gains*

The ability to selectively recruit the appropriate number of actuators for a given force requirement is a novel actuation method that provides energy savings for pneumatic and hydraulic systems. The efficiency gains that come from a variable recruitment actuation scheme have been shown through modeling and offline recruitment experiments [10], [16], [17], [18]. The following quasi-static analysis is shown to demonstrate the efficiency gains that can come from variable recruitment. This study makes use of the Tondu McKibben muscle model found in equation (1) to simulate the energy savings of an idealized variable recruitment muscle bundle actuating against a linear spring. The muscle bundle for this example was assumed to have six McKibben muscles arranged in a parallel configuration and activated in pairs. The braid angle of the muscles was set to 29 degrees.

Fig. 6(a) shows the force-stroke curve of the muscle bundle with two, four, and all six muscles activated. The spring load curve is also plotted in Fig. 6(a). All force values were normalized by the maximum load the fully-activated muscle bundle could produce at a given pressure for a given muscle geometry. It is important to notice that actuating against a linear spring is one of the most inefficient loading scenarios for McKibben muscles, as the spring force-strain curve is approximately the inverse of the muscle force-strain curve. Therefore, as the muscles start to pull against the spring, the force required to displace the spring quickly builds and more muscles need to be recruited. These recruitment thresholds are shown as points A and B in Fig. 6.

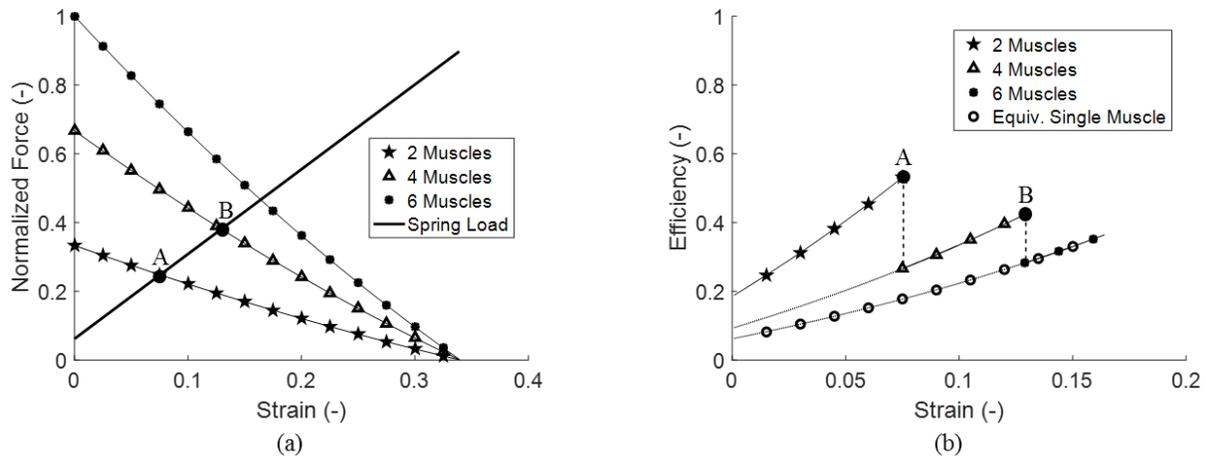

Fig. 6. a) Normalized force-strain plot of a muscle bundle containing six McKibben muscles with variable recruitment actuating against a linear spring. b) The corresponding efficiency curves with the equivalent single muscle case without variable recruitment included for comparison. Points A and B indicate recruitment thresholds.

The efficiency curves as a function of strain for two, four, and all six muscles are shown in Fig. 6(b). As stated previously, the size of a hydraulic actuator is traditionally determined by the maximum force output needed for a particular application. Therefore, for this example, the traditional approach would be to use one actuator with an equivalent cross-sectional area to the total cross-sectional area of the six muscles in the bundle to meet the required force output. This equivalent single actuator efficiency is also shown in Fig. 6(b). The source of the efficiency gain can be seen in this figure. Initially, two muscles are active and pull against the spring until the force requirement becomes too great for the two muscles. Then, at point A, two additional muscles are recruited. This causes the drop in efficiency shown in Fig. 6(b) due to the sudden increase in actuator area, which causes a decrease in muscle pressure and additional throttling losses. This is then repeated at point B, when the force requirement becomes too great for the four muscles. At this point, two additional muscles are recruited and the efficiency drops to follow the equivalent single actuator curve (which is the efficiency curve that traditional hydraulics would follow). In general, variable recruitment provides energy savings by selectively activating the number of muscles needed to meet the instantaneous force requirement, which reduces throttling losses by allowing the muscle pressure to remain closer to the operating pressure of the system.

## *Testing and Discussion of the LHACD and McKibben Muscles*

Due to the versatility of the LHACD, a wide range of tests were performed to validate this test bed, characterize the artificial muscles, and explore the power savings of a variable recruitment control scheme. First, we performed several tests on the LHACD to characterize its dynamic behavior. Next, we performed characterization tests on our McKibben artificial muscles to determine their quasi-static force-stroke capabilities. Finally, we operated the artificial muscles with and without variable recruitment during identical loading scenarios to demonstrate the power savings associated with variable recruitment.

*LHACD Characterization Tests*

The dynamic capabilities of the LHACD were determined by testing the drive cylinder bandwidth, maximum velocity, and response to a step input.

First, tests to characterize the bandwidth of the drive cylinder on the LHACD were completed. For these tests, the unloaded drive cylinder attempted to track a sine wave with an amplitude of 8 inches at varying frequencies at an operating pressure of 600 psi. At an amplitude of 2 inches, which is a typical stroke for the meso-scale applications in which we are interested, a frequency of approximately 1.4 Hz was achieved. A standard human walking gait has a frequency of 0.86 Hz and a running gait of 1.33 Hz, therefore, the ability of the LHACD to meet the typical stroke requirement for legged robots at a frequency of 1.4 Hz proves to be adequate for this application [29], [30]. The complete results from these tests are shown in Fig. 7, as well as the bandwidth values of commercially available universal testing machines. In comparison to these commercial universal testing machines, the LHACD offers a unique combination of speed and crosshead travel that is not seen in other devices.

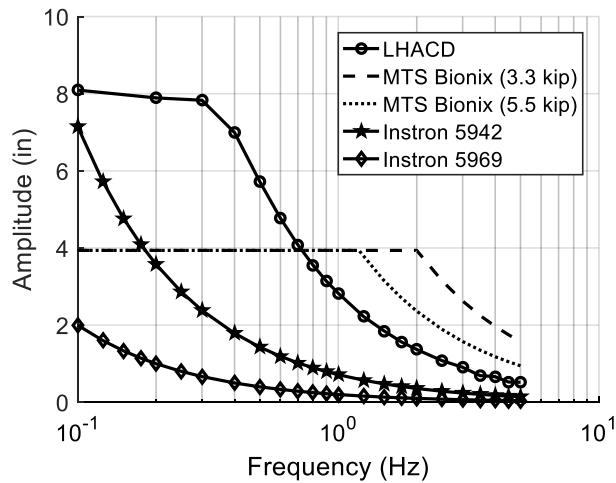

Fig. 7. Bandwidth testing results for the drive cylinder of the LHACD compared with bandwidth values of other commercial devices (Instron 5942 and Instron 5969 bandwidth values were estimated based on the given maximum extension and retraction speeds of the device).

Next, we performed a series of tests on the LHACD to characterize the open loop gain for the drive actuator. For these tests, a constant voltage signal was sent to the drive cylinder's servo valve, and then the velocity of the unloaded cylinder rod was measured. The complete data set for the full range of input voltages is shown in Fig. 8. As seen in this figure, the velocity of the cylinder rod at an input voltage of zero volts is slightly negative. This is a safety feature to ensure that the cylinder rod gently retracts in case the servo valve does not receive a signal.

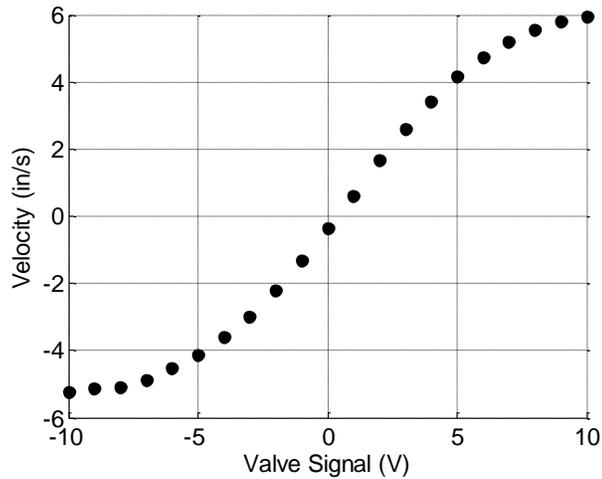

Fig. 8. Open loop gain test results for the drive cylinder of the LHACD.

Finally, the response of the LHACD drive cylinder to a step input was measured. For this test, the drive cylinder began at its minimum stroke and received a step input to track a reference stroke of 10 inches. The step response is shown in Fig. 9. During this test, the applied valve voltage signal saturated at 10 volts. Therefore, from this data, the maximum velocity of the drive cylinder was determined, as well as the system's time constant. These characteristics, as well as others, are shown in Table 2. These attributes exemplify the versatility of the LHACD and demonstrate the wide range of tests that can be performed on linear actuators of various design and size.

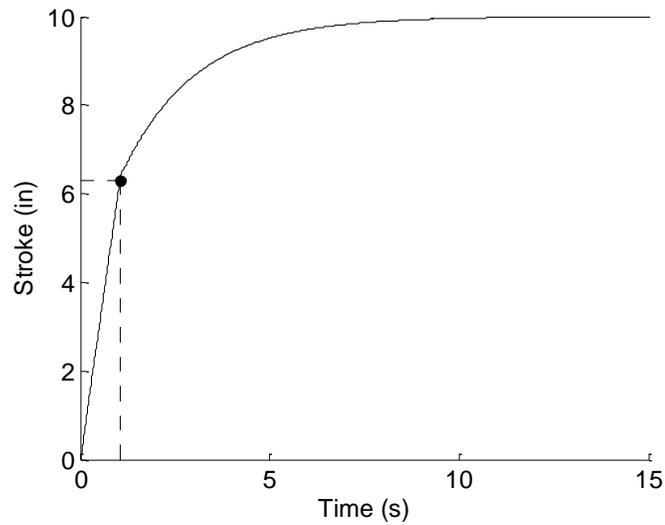

Fig. 9. Response of the LHACD to a step input of 10 inches. The vertical dashed line represents the time constant for the LHACD.

Table 2. Characteristics of the LHACD.

| | |
|---|---|
| Maximum Force | 2700 lbf (extension) * |
| | 2200 lbf (retraction) * |
| Maximum Stroke of Drive Cylinder | 12 in. |
| Maximum Stroke of Muscles | 7 in. (LHACD can accommodate muscles up to 24 in. in length) |
| Bandwidth | 1.4 Hz (at 2 in. amplitude) |
| Maximum Velocity | 7.84 in/s (extension) ** |
| | 9.34 in/s (retraction) ** |
| Time Constant | 1.1 s (for a step input of +10 in.) |

\* Ideal calculated values based on cylinder dimensions and theoretical stalled condition
\*\* Ideal calculated values based on cylinder dimensions and no-load condition

In the current state of research, linear fluid power actuators – and especially pneumatic McKibben muscles – are commonly tested against a spring or static weight, as described previously. However, another common testing method makes use of a universal testing machine, such as an Instron or MTS machine [28]. To compare the capabilities of the LHACD to other available universal testing machines on the market, Fig. 7 and Table 3 display the key attributes of the devices [31], [32]. Table 3 shows that the LHACD offers a unique combination of force, speed, and travel, which allows it to perform dynamic tests using a wide variety of user-specified stroke and force profiles that are relevant to bio-inspired robotics actuation studies. While the Instron machines offer large crosshead travel, their speeds are much lower than the LHACD. Conversely, the MTS Bionix systems produce comparable or greater speeds than the LHACD, but much less travel. Most importantly for this research, however, the LHACD is fully programmable to control both the drive cylinder and the variable recruitment test actuators, and instrumented to simultaneously collect force, displacement, applied pressure, and flowrate characterization data.

Table 3. Comparison of the LHACD and various universal testing machines

|  | LHACD | Instron 5942 | Instron 5969 | Instron 5989 | MTS Bionix 370.02 (3.3 kip) | MTS Bionix 370.02 (5.5 kip) |
|---|---|---|---|---|---|---|
| Extension Load Capacity (lbf) | 2,700 | 112.5 | 11,250 | 134,800 | 3,300 | 5,500 |
| Retraction Load Capacity (lbf) | 2,200 | 112.5 | 11,250 | 134,800 | 3,300 | 5,500 |
| Extension Speed (in/s) | 9.34 | 1.67 | 0.4 | 0.33 | 9.89 * | 5.94 * |
| Retraction Speed (in/s) | 7.84 | 1.25 | 0.4 | 0.33 | 9.89 * | 5.94 * |
| Crosshead Travel (in) | 12 | 19.2 | 44.9 | 72.8 | 4 | 6 |
| Vertical Test Space | 32 | 28.6 | 47.7 | 78.8 | 32.6 | 32.6 |
| Dimensions (height x width x depth) (in) | 68 x 12 x 6 | 39 x 18 x 24 | 64 x 31 x 29 | 123 x 63 x 38 | 78.3 x 24.5 x 22.7 | 78.3 x 24.5 x 22.7 |

* Estimated values based off of Displacement-Frequency bode plot from MTS Bionix datasheet [32]

Once we quantified the capabilities of the LHACD, we then conducted tests to determine the quasi-static force-stroke capabilities, as well as the hysteretic behavior, of our hydraulic artificial muscles at various pressures. For each test, the pressure in the single muscle was held constant. This was achieved through the use of a PID controller, with the muscle's internal pressure as the controlled variable. Then, with the McKibben muscle initially at zero percent strain, the drive cylinder slowly extended until the muscle was at full contraction, and then the drive cylinder slowly retracted until the muscle was again at its full resting length. Meanwhile, the force generated by the artificial muscle at the varying stroke lengths was recorded. The drive cylinder was controlled by a PID controller, as described previously, and was made to track a low frequency sinusoidal stroke trajectory. The results from these tests are shown in Fig. 10. From this figure, we can see the force that a single artificial muscle can produce at a given pressure, the maximum force that it can produce, as well as its maximum strain.

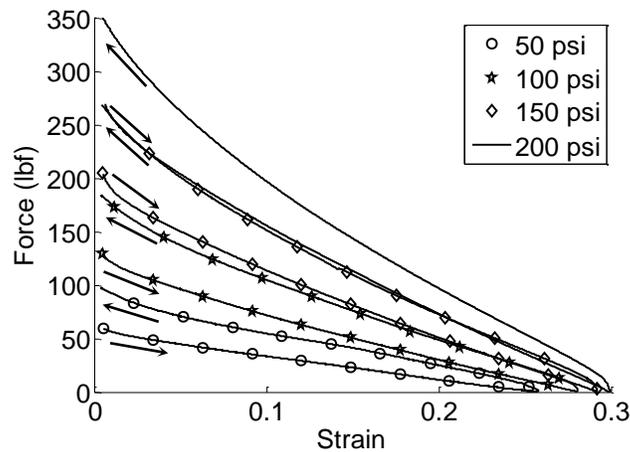

Fig. 10. Quasi-static force-stroke capabilities of our in-house constructed artificial muscles at varying supply pressures.

*Efficiency Tests for Variable Recruitment*

Here we present a first measurement of dynamic trajectory tracking with online variable recruitment to quantify the energetic benefits of variable recruitment in a hydraulic artificial muscle actuation system. For these tests, a 0.2 Hz sine wave with a peak-to-peak amplitude of 1 inch was prescribed as the position trajectory to the drive actuator. Simultaneously, a synchronized 0.2 Hz sine wave with a peak-to-peak amplitude of 500 lbf was prescribed as the force trajectory to the artificial muscle bundle. These input trajectories were biased such that the muscles contracted from 0 to 1 inch (10% strain) at 0.2 Hz, while simultaneously experiencing a load oscillating from 100 lbf to 600 lbf. With these same force and stroke curves prescribed to both the variable recruitment and fixed recruitment cases, the power consumption with and without variable recruitment can be experimentally compared. The stroke and force curves for the cases with and without variable recruitment, as well as the corresponding flow rates, are shown in Fig. 11.

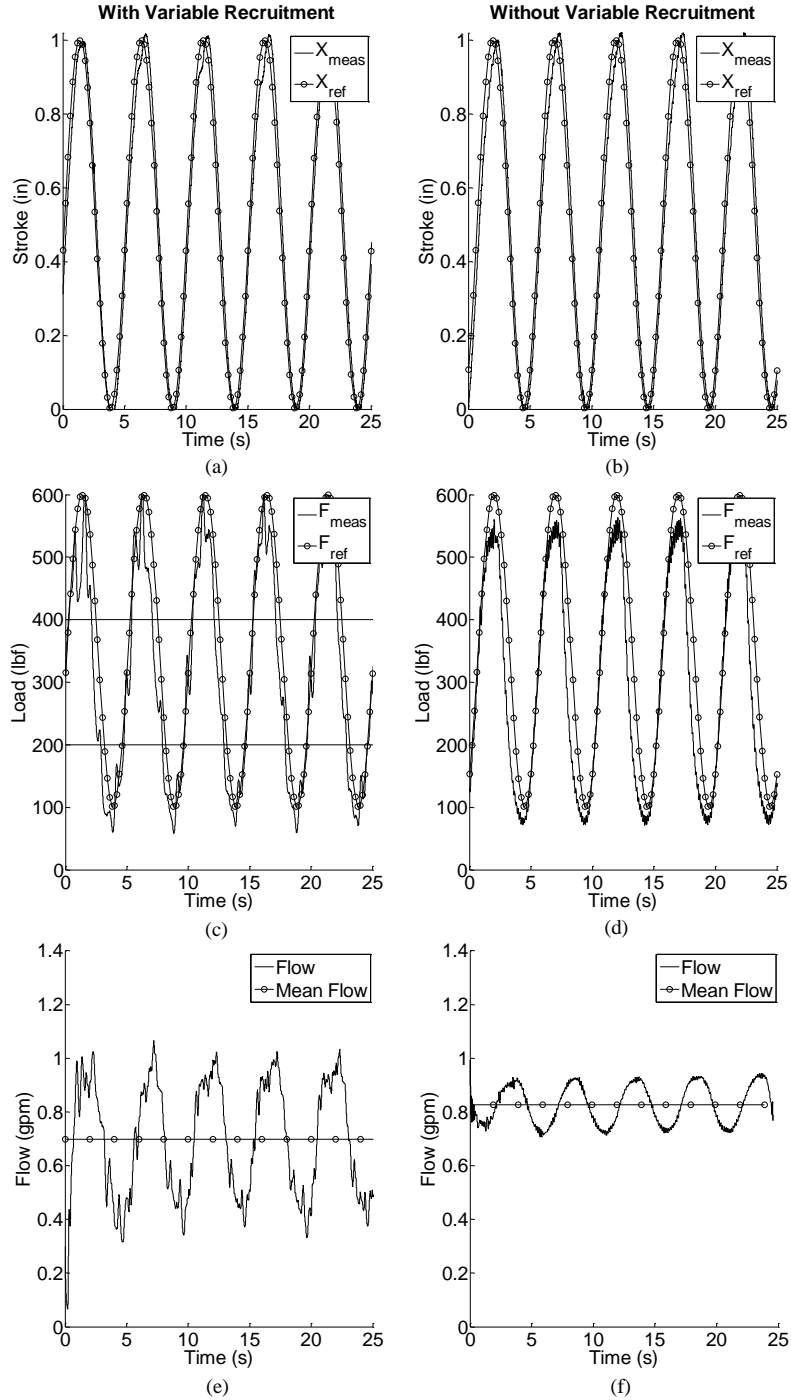

Fig. 11. a) Measured and prescribed stroke profiles with variable recruitment. b) Measured and prescribed stroke profiles without variable recruitment. c) Measured and prescribed force profiles with variable recruitment (horizontal lines indicate recruitment thresholds). d) Measured and prescribed force profiles without variable recruitment. e) Measured flow rate with variable recruitment. f) Measured flow rate without variable recruitment.

For the tested variable recruitment actuation scheme, the first muscle pair was always active, the second muscle pair became active after a reference force level of 200 lbf was exceeded, and finally, the third muscle pair became active once a reference force level of 400 lbf was exceeded. This actuation scheme also incorporated down-recruitment (i.e. a muscle pair became inactive once the force level dropped below the specified threshold). These force thresholds at which another muscle pair was either activated or inactivated are shown by the horizontal lines in Fig. 11(c).

As seen in Fig. 11(e) and 11(f), the average flow to the McKibben muscles was reduced for the test with variable recruitment, compared to the case without variable recruitment. In both cases, the force and stroke trajectories were followed reasonably well. Therefore, the mechanical power output for both tests were similar. However, due to the decrease in flow delivered to the muscles during the test with variable recruitment, the fluid power input was also reduced. Therefore, the test with variable recruitment yielded a higher efficiency than the case without variable recruitment. A summary of the results from this test are shown in Table 4.

Table 4. Power savings of a variable recruitment actuation scheme

|  | Average Fluid Power In | Average Mechanical Power Out | Average Efficiency |
|---|---|---|---|
| Without Variable Recruitment | 16.67 W | 7.89 W | 0.473 |
| With Variable Recruitment | 13.84 W | 7.91 W | 0.571 |
| Percent Improvement | -16.97% | +0.21% | +20.7% |

Further controller development that addresses the highly non-linear behavior of the McKibben artificial muscles will allow for a wider range of force and stroke trajectories to be simultaneously prescribed to the test actuators for efficiency testing. Furthermore, other complexities involved with controlling artificial muscles using a variable recruitment actuation scheme can also be addressed through further controller development, such as the minor oscillations seen in Fig. 11(c) around the switching thresholds. In spite of the simplistic controller implemented here, the results demonstrate that online variable recruitment of fluidic artificial muscles can deliver significant energy efficiency gains of over 20 percent (0.098 higher) while tracking a time-varying trajectory, and thus provide an experimental proof-of-concept for this novel actuation approach.

### *Conclusions Drawn from Hydraulic Actuation Study*

In an effort to improve the actuation efficiency for meso-scale hydraulic systems, we developed the LHACD to investigate hydraulic artificial muscle bundles using a variable recruitment actuation and control scheme. The LHACD was designed and constructed to provide a versatile and convenient experimental apparatus on which hydraulic actuators can be tested under different loading scenarios and characterization data can be seamlessly collected. The hydraulic actuators of interest for this study are McKibben artificial muscles, which were developed and fabricated in-house. This study validated the LHACD by testing its capabilities and comparing its characteristics to similar devices. Furthermore, the LHACD allowed for characterization tests of the hydraulic artificial muscles to be performed by measuring the quasi-static force-stroke capabilities. Finally, an online variable recruitment control scheme was implemented and tested for a bundle of hydraulic McKibben muscles, a novel test procedure that has been lacking from the research community. This variable recruitment control approach, while simplistic in nature for this study, demonstrated a significant reduction in fluidic power consumption while tracking an oscillatory force and displacement trajectory. At the same time, the variable recruitment bundle produced nearly identical mechanical power output to an equivalent case without variable recruitment, leading to an efficiency gain of over 20 percent (0.098 higher) in the scenario tested. Future studies will incorporate a more advanced control scheme that will address the highly non-linear behavior of McKibben muscles. This will allow for future tests to be performed to further measure and optimize the efficiency gains associated with a variable recruitment actuation scheme. Due to the versatility and convenience that the LHACD offers for actuator testing and controller development, these and related tests can be extended to a wide range of hydraulic actuators for a myriad of applications in robotics, industrial automation, or aerospace. Thus, this testing approach provides hardware-in the-loop energetic quantification and dynamic validation that a given hydraulic actuator is able to meet the requirements for a desired application.

CHAPTER 2

AUTONOMOUS URBAN LOCALIZATION AND NAVIGATION WITH LIMITED

INFORMATION

*Introduction to Autonomous Urban Navigation*

Autonomous driving, as with many other robotic applications, requires accurate localization to perform robustly. However, dense urban environments provide a key challenge due to the lack of reliable information sources for localization and planning. For example, multipath errors and signal shadowing in dense urban environments make positioning systems based on GPS an unreliable information source for autonomous agents [33]. Even highly accurate, state-of-the-art positioning systems struggle to provide the level of localization needed for autonomous driving due to the difficulties that dense urban environments present [34]. Additionally, highly detailed maps typically needed for autonomous driving are highly sensitive to environmental changes, and are expensive to obtain in regard to time, money, and resources. Storing these high-fidelity maps on-board the vehicle is unrealistic for all maps in all locations or for environments that change, such as construction areas often found in cities. Data connections could be relied upon to provide the vehicle access to highly detailed maps, however, these connections can be weak, spotty, or non-existent in urban areas. Furthermore, security of autonomous vehicles is also a major concern, as both GPS measurements and maps can be spoofed and/or jammed [35]. Given these challenges, this study investigates alternative architectures and sources of information that may be used for robust navigation of urban roadways.

Currently, most autonomous driving systems use high precision GPS signals to localize within a highly detailed environmental map. Many of the foremost competitors in the autonomous driving industry, such as Google, Uber, and Ford, have entire teams dedicated to obtaining and updating their high-definition (HD) maps, which include everything from lane markings to potholes. These maps are acquired by manually driving vehicles while collecting 360-degree lidar and/or camera data of the environment in which the autonomous vehicle will later drive [36]. This data then undergoes heavy post-processing to form the HD map. Due to the extreme complexity of this task, automakers such as Volkswagen, BMW, and General Motors have relied on third-party services, such as HERE and MobilEye, to provide these highly detailed maps.

To avoid the difficulty of obtaining HD maps in which to localize the vehicle using high precision GPS, research efforts exist in the robotics community to address the problem of navigating without such high-fidelity information sources. On-line *Simultaneous Localization and Mapping* (SLAM) can be applied to autonomous driving to alleviate the need for precise GPS measurements and highly-detailed maps [37], [38]. For example, FAB-MAP, a topological SLAM technique, has been shown to allow for appearance-based navigation [39]. Similarly, SeqSLAM is another SLAM technique that aims to allow for visual navigation despite changing environmental conditions [40]. Although both methods do not rely on GPS measurements or an *a priori* map of the environment, they navigate purely in appearance space and make no attempt to track the vehicle in metric coordinates; in other words, the techniques behave similar to a scene-matching algorithm. Without a way to track the vehicle in metric coordinates, it is impossible to locate the vehicle when it is in between two matched scenes. Therefore, a sufficiently-dense map of images is needed for adequate localization, which causes the algorithms to quickly become computationally expensive and offer sharply diminishing performance as the scale of the environment grows.

Techniques which are less computationally arduous can address the problems of navigating without GPS and a highly detailed map separately. To reduce the vehicle's reliance upon GPS, several methods have been proposed in recent years based on accurate self-localization in mapped environments [41], [42], [43], [44], [45]. However, these techniques still rely heavily on *a priori* map information, coming either from lidar or vision data. Similarly, a PosteriorPose algorithm has been shown to keep the navigation solution converged in extended GPS blackouts by augmenting GPS and an inertial navigation system with vision-based measurements of nearby lanes and stop lines referenced to a known map of environmental features [46]. This algorithm retained a converged and accurate position estimate during an 8-minute GPS blackout. To address autonomous navigation without *a priori* map data, GPS-fused SLAM techniques have also been proposed [47], [48]. However, the assumption of consistently receiving these GPS measurement updates is not valid for urban applications, such as in urban canyons like Manhattan and Chicago, and therefore should not be relied upon.

While recent research efforts have been made to face the challenges of driving either in GPS-denied circumstances or without an HD map of the environment, there is considerably less research to simultaneously address both challenges in a computationally-efficient manner. This work aims to explore the extent to which a vehicle can navigate in an urban environment while assuming a varying degree of external information. First, as a worst-case situation, this work studies how far a vehicle can travel with no GPS measurements and no *a priori* map information, other than an initial starting location and measurements from wheel encoders and a compass. Second, this chapter explores how far a vehicle can travel with a minimal amount of information, which, for the purposes of this study, comes as a sparse map of landmarks. This results in a navigation solution with much better scalability than many other techniques in the research community. While the focus of this chapter is on navigation with limited information, this work could also be used to supplement current navigation systems that use GPS, HD maps, and/or SLAM techniques. This supplemental technique could provide indispensable aid to autonomous vehicles for cases when the navigation system experiences difficulties or its security is threatened; this robustness to infrequent, yet crucial, events is critical for a long-term navigation solution.

### *Autonomous Navigation System Architecture*

*Autonomous Navigation System Overview*

The system architecture developed for this study is shown in Fig. 12. This system diagram contains common elements to autonomous driving such as steering and speed controllers, an object tracker, and a path generator. However, the pose estimator and navigation algorithm are updated from their typical form to address the challenges associated with the lack of map information and GPS measurements. The proposed algorithm assumes local sensors allow for the vehicle's real-time control (i.e. staying in a lane). Therefore, precise in-lane localization is not needed for this approach. Rather, high-level localization is provided by the pose estimator, which utilizes only odometry measurements, compass measurements, and sparse map-based measurements, which come from an on-board sparse map of landmarks with corresponding coordinates. This estimator is termed "lightweight" due to the limited amount of sensor measurements it requires. The sparse map-based measurements generated from computer vision methods compare raw camera images to landmark images contained within a sparse map.

For the purposes of this study, the map information is assumed to be limited (i.e. no global roadmap, but only a sparse map of images and their corresponding set of coordinates).

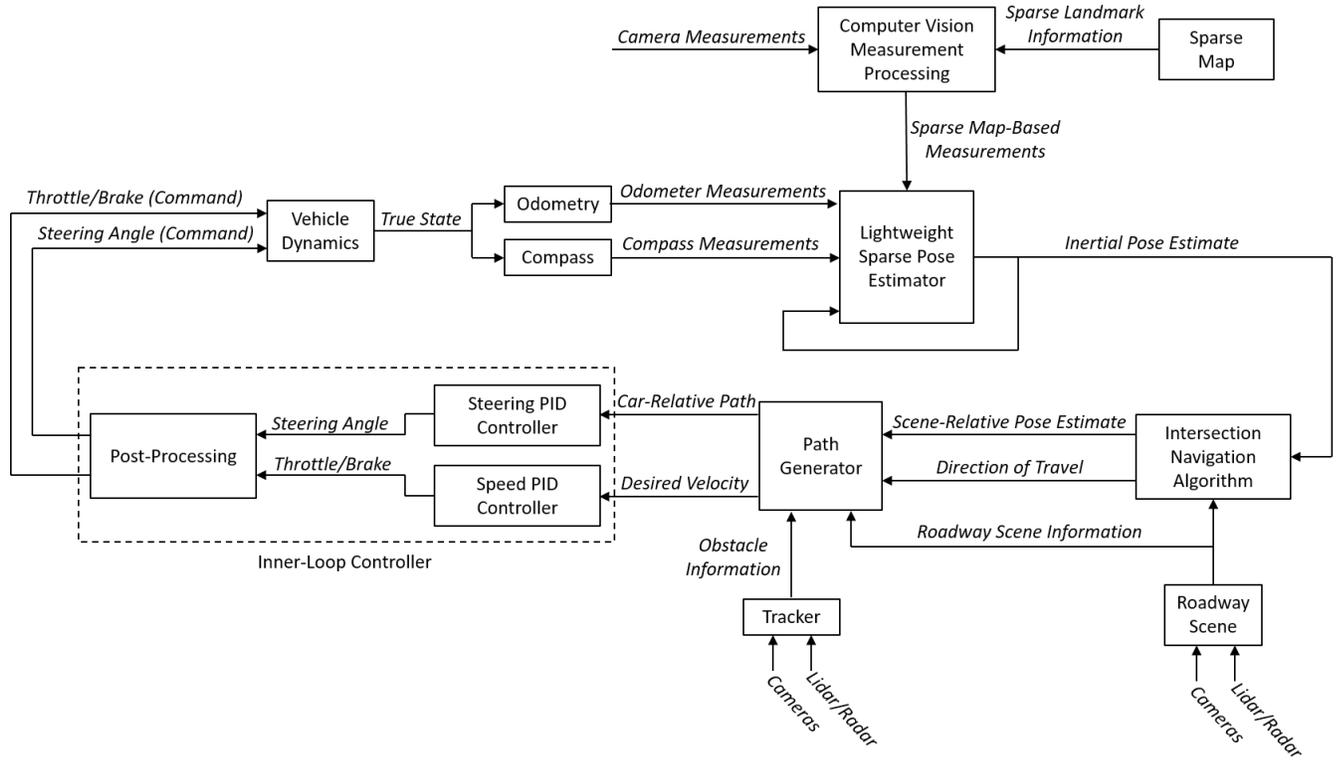

Fig. 12. System architecture for autonomous urban navigation with limited external information.

The roadway scene includes information such as lane line markings, road signs, traffic lights, and other roadway information that can be extracted from sensor measurements. However due to the focus of this study on navigation and estimation, the roadway scene is assumed to be known. Finally, the roadway scene information, along with the inertial pose estimate, feeds into an intersection navigation algorithm and is used to probabilistically determine the best route to take to reach the goal based on the current limited belief. This high-level navigation scheme is provided by a compass-based navigation control law.

*Lightweight Sparse Pose Estimation*

As a first step in this study, a pose estimator is developed assuming no detailed *a priori* map information and no GPS measurements. The pose estimator relies on globally-known start and goal locations of the vehicle, dead reckoning using odometry and compass measurements to estimate the pose of the vehicle within a local frame of reference, and measurement updates to sparse, but known, landmarks. Dead reckoning using odometry measurements alone diverges over time. However, the addition of a compass measurement update, which directly measures the heading of the car, improves the pose estimator accuracy notwithstanding relatively high compass uncertainty. To enable longer driving, the pose estimator is improved using sparse map-based measurement updates. This allows two key questions to be addressed in this work: 1) How far can the vehicle travel during a GPS blackout and with no map information? 2) What level of map landmark sparsity enables the vehicle to successfully navigate a certain distance?

A sparse map of landmarks with corresponding images and coordinates is assumed to be contained within a database on the vehicle. Therefore, as the vehicle travels in the environment and collects camera data, it performs scene detection via computer vision techniques to compare the collected images to the images of landmarks within its database. In this work, ORB feature detection is used to detect and describe features within the local scene of the vehicle and then match to a corresponding urban scene within the database of images on the vehicle reference [49]. When the ORB feature detector obtains a test image that matches an image in the landmark database, the corresponding landmark location is used as a measurement update within the extended Kalman filter (EKF) to refine the pose estimate of the vehicle.

This technique is implemented in simulation as described in the simulation results section of this chapter, and then in experiments as described in the experimental results section of this chapter. It is noted that the proposed work can make use of any scene detector or computer vision method without loss of generality. However, ORB feature detection was chosen rather than a more sophisticated approach, such as the bag-of-words model used in FAB-MAP, due to its ease of implementation and the fact that it does not need to be extensively trained and tuned. Furthermore, by tracking the vehicle's position in metric coordinates in between landmark detections via dead reckoning, a mask can be generated to disregard all landmarks beyond the 2-sigma uncertainty ellipse around the vehicle. This caused the ORB landmark detector to be very computationally efficient – as it did not need to compare the current test image to all the images in the database, but only to the database images closest to the vehicle – and it also improved performance, as it essentially eliminated all false positives for the experiments described later in this chapter.

*Intersection Navigation Algorithm*

Navigating to a desired location in an urban environment without the availability of GPS or detailed map information is a difficult challenge. Certain assumptions must be made for this problem to become feasible. First, the coordinates of the start and end points are assumed to be known within a defined uncertainty. Next, the vehicle is assumed to be equipped with a suite of sensors that allows it to detect a variety of common roadway objects such as road signs, traffic lights, cyclists, pedestrians, other vehicles, and lane markings. The ability to detect at this level is important because it allows the vehicle to avoid obstacles, stay on the road by following lane lines, and detect when it is approaching an intersection; however, these measurements are not used for pose estimation. Given the readily available lidar units, radar units, cameras, and computer vision technology on the market, this is a valid assumption to make. For instance,

MobilEye's vision-based advanced driver assistance system can detect lane lines, other vehicles, pedestrians, and various traffic signs [50]. Many autonomous vehicles have this detection capability. However, since this study aims to address novel pose estimation and navigation techniques, these detection capabilities were not deemed necessary to implement for this research.

Assuming such a suite of sensors is available, the low-level navigation problem becomes feasible, and a higher-level navigation problem can be formulated. Navigation to a desired location from a given starting point follows two basic principles in this framework. First, as the vehicle approaches an intersection and is faced with a decision of its direction of travel (e.g. continue straight, turn left, turn right, etc.), the vehicle minimizes the difference in angle between the heading of the vehicle and the direction to the goal after the proposed intersection decision. Second, the vehicle retains a list of intersections that it has already visited and the corresponding decision made at each intersection; note that this list of visited intersections is based on the vehicle's pose at the time an intersection is detected, which is subject to uncertainty and will become less reliable as the pose uncertainty increases. As the vehicle approaches a previously-visited intersection, the navigation algorithm applies a penalty to the prior intersection decision, thus, making it less probable to make the same decision as before. The second principle is similar to Tabu search [51], as it relaxes the basic rule of the algorithm and discourages the search to return to previously-visited solutions to avoid getting stuck in suboptimal regions. The decision rule is general to any type of intersection (e.g. different number of roads at different angles), although this study focuses on gridded roadways due to the intended application to urban environments.

These two principles for the navigation algorithm are summarized in the following optimization problem:

$$\theta^* = \underset{i \in N}{\operatorname{argmin}}[|\theta_i^+ - \varphi_i| + \gamma_i] \tag{4}$$

where $\theta^*$ is the intersection decision, or direction of travel after the intersection, $i$ represents the index of turning options at an intersection (e.g. continue straight, turn left, turn right, etc.), and $N$ is the total number of turning options at a given intersection. The direction of travel of the vehicle after the intersection is denoted by $\theta_i^+$. The direction from the vehicle to the goal after the intersection is represented by $\varphi_i$. The penalty of making the same decision at the same intersection as before is given by $\gamma_i$, which increases incrementally each time the same decision is made at the same intersection.

Fig. 13 shows an example of an intersection decision to better understand the navigation algorithm in (4). In this example, the vehicle navigates through a 3-way intersection. The navigation algorithm assumes that the vehicle can detect the type of intersection that it is approaching (e.g. 3-way intersection) based on the measurements of its on-board sensors. With this information, the algorithm computes the angle between the vehicle's heading and the direction to the goal for each possible route. After including any penalties for the decisions made at prior visits, the route associated with the minimum value is selected. In this example, assuming no penalties, the algorithm would direct the vehicle to go straight through the intersection.

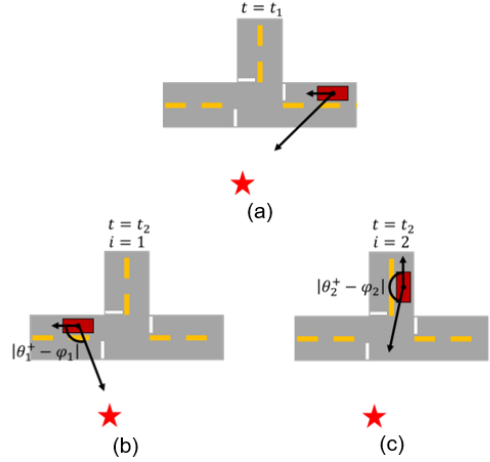

Fig. 13. Illustration of an autonomous vehicle navigating through a 3-way intersection. The goal is denoted by the red star. (a) The vehicle approaching the intersection. (b) The vehicle after continuing straight through the intersection (decision $i = 1$). (c) The vehicle after turning right through the intersection (decision $i = 2$). The navigation algorithm would choose to continue straight through the intersection in (b), assuming no previous visits to this intersection.

*Simulation Results for Autonomous Navigation with Limited Information*

A simulator was developed to test the feasibility of the proposed pose estimation and navigation techniques. The simulator models the vehicle's dynamics using a four-state bicycle model [52] and simulates its motion through a randomly-generated city road network. A predictive state model of the vehicle is given as

$$\bm{x}_{k+1} = f(\bm{x}_k, a_k, \varphi_k, \delta t) \tag{5}$$

$$\begin{bmatrix} x_{k+1} \\ y_{k+1} \\ \theta_{k+1} \\ v_{k+1} \end{bmatrix} = \begin{bmatrix} x_k + \Delta x \cos(\theta_k) - \Delta y \sin(\theta_k) \\ y_k + \Delta x \sin(\theta_k) + \Delta y \cos(\theta_k) \\ \theta_k + \Delta \theta \\ v_k + a_k \delta t \end{bmatrix} \tag{6}$$

In the equations above, $\bm{x}_k$ is the vehicle state at time $k$, which consists of four states: the position of the center of the vehicle's rear axle ($x_k$ and $y_k$), the heading of the vehicle ($\theta_k$), and the speed of the vehicle ($v_k$). The control inputs to the vehicle model are acceleration ($a_k$) and steering angle ($\varphi_k$). The prediction time step is given as $\delta t$. In addition, the $\Delta$ terms are defined as

$$\Delta\theta = \frac{d_k}{\rho_k}, \quad \Delta x = \rho_k \sin(\Delta\theta)), \quad \Delta y = \rho_k(1 - \cos(\Delta\theta))$$,

where

$$\rho_k = \frac{l}{\varphi_k}, \quad d_k = \frac{1}{2}a_k \delta t^2 + v_k \delta t$$.

The parameter $l$ is the length between the front and rear axles, $\rho_k$ represents the radius of curvature for the vehicle, and $d_k$ represents the distance the vehicle travels over $\delta t$.

During the vehicle's travel, simulated odometry, compass, and vision measurements are used to update the vehicle pose estimate. Steering and speed proportional-integral-derivative (PID) controllers are used to allow the vehicle to stay on a specified path chosen by the navigation algorithm. A randomly gridded map was generated for each test in this Monte Carlo study. These maps ranged in size from 1 sq. km to 100 sq. km, and the block size ranged from 50 meters to 300 meters. Dead end and one-way roads were randomly scattered into the map. Random start and end locations were chosen. Finally, for the landmark detection study, landmarks were scattered into the map at random locations according to the given landmark density.

*Monte Carlo Range Tests without Map Information*

Monte Carlo studies were conducted in simulation to determine the maximum distance an autonomous vehicle can travel without receiving external pose measurements (i.e. no GPS or landmark measurements) before it becomes lost. Without the use of a map, local sensors are assumed to enable the vehicle to stay on the road. The limiting factor in reaching the goal is the navigation algorithm's ability to differentiate between each intersection. At a sufficiently high position uncertainty level, it becomes ambiguous which intersection the vehicle is approaching and the high-level navigation breaks down. The vehicle is assumed to be lost when the major axis of the 2-sigma uncertainty ellipse of the vehicle's x-y location grows to exceed 100 meters, which is approximately the average size of a Manhattan city block [53]. A 2-sigma uncertainty ellipse, rather than a 1-sigma uncertainty ellipse, is used to provide a high confidence region for the location of the vehicle. While the uncertainty is below this threshold, the vehicle is considered sufficiently localized for the high-level navigation algorithm to function.

This study assumes a conservative odometer measurement uncertainty common in cars and ABS braking systems. This uncertainty considers wheel slip and rotary encoder discretization errors, which have a maximum error of ±1/2 of the angular rotation between two successive bits [54]. For compass measurements, interference from the large amount of electrical hardware found on an autonomous vehicle can result in high uncertainty. Therefore, three values for the 2-sigma compass uncertainty are studied here: ±10°, ±20°, and ±30°. For comparison, an additional set of simulations were performed in which no compass measurements were used. A total of 4,000 simulations were conducted, with each simulation using a new random map, and new random start and end points. Fig. 14 plots the 2-sigma ellipse major axis as a function of distance traveled for each trial in all three compass uncertainty cases, as well as the case with no compass measurements.

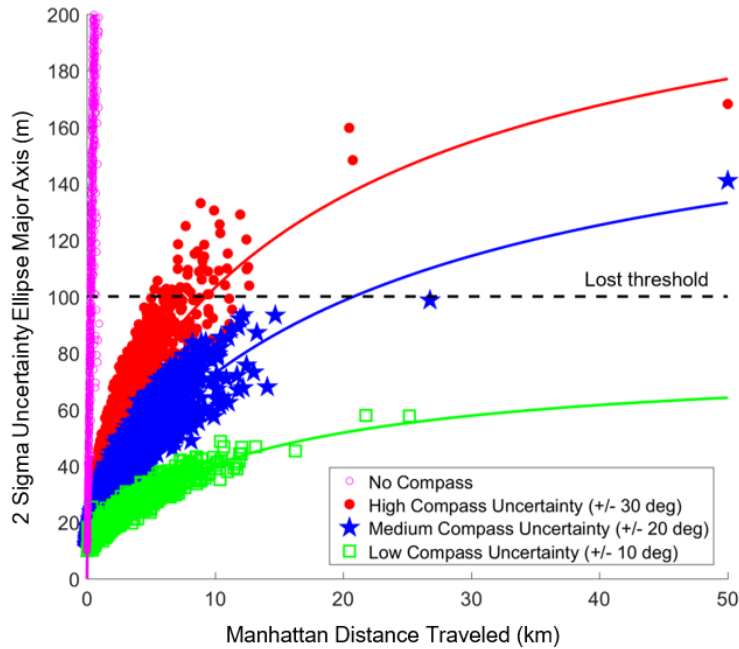

Fig. 14. Monte Carlo results to determine the distance an autonomous vehicle can travel without external pose measurement updates before getting lost, which is defined by the threshold indicated by the horizontal dashed line; note that this figure shows the Manhattan distance traveled by the car.

The results for each set of tests follow a predictable trend: the major axis of the 2-sigma uncertainty ellipse increases quickly when the ellipse is small and then grows more gradually as the area of the ellipse becomes large. For a 2-sigma compass uncertainty of ±10°, a distance at which the vehicle becomes lost was not found. For a 2-sigma compass uncertainty of ±20°, the vehicle could travel 20.8 km on average before it became lost. For a 2-sigma compass uncertainty of ±30°, the vehicle could travel 9.3 km on average before it became lost. Finally, for the case with no compass measurements, the vehicle could only travel 300 meters on average before it became lost. In this case, the 2-sigma uncertainty grew very rapidly since only wheel encoders were being used to estimate the position of the car. Therefore, as the vehicle began to drive, the initial uncertainty in the heading of the vehicle quickly resulted in a large amount of lateral uncertainty in the position of the vehicle.

The main source of variation in this study relates to the number of turns that the vehicle made during its travel. It is expected that the final uncertainty ellipse major axis is larger for a vehicle driving along a straight road compared to a vehicle driving the same distance while taking many turns. This is seen in Fig. 14, as most of the data points above the fitted curve resulted from tests in which the vehicle took few turns, while the majority of the data points below the fitted curve resulted from tests in which the vehicle took many turns. This is because the ellipse grows predominantly along only one axis in the case of driving straight and along both axes in the case involving many turns, and therefore the major axis of the 2-sigma uncertainty ellipse grows at a faster rate for straight driving.

*Monte Carlo Landmark Detection Study*

A Monte Carlo study was performed to determine the effect that map-based measurements from sparse landmarks have on the pose estimation problem and subsequent navigation. The parameters of this study were the same as the prior Monte Carlo range tests, except this study utilizes a sparse map of landmarks with corresponding coordinates known *a priori* to allow the vehicle to perform pose updates as it drives. A camera takes measurements of a scene, and attempts to correlate a detection with a sparse map of locations. If a positive detection is made, a map-based pose measurement update is performed using an uncertain location.

Furthermore, this study explores three variants of the heading-based navigation function in (4). The first method, termed the straight to goal method, attempts to drive straight to the goal and does not actively seek out landmarks (i.e. fortuitous landmark measurement updates). The second method, called the landmark to landmark method, seeks out the closest landmark to the vehicle that also moves the vehicle closer to the goal. The third approach is the hybrid method, where the vehicle attempts to drive straight to the goal until its pose uncertainty exceeds a specified threshold, at which point the vehicle then seeks out landmarks to improve its pose estimate. For this study, the specified threshold is 50 meters for the major axis of the 2-sigma position uncertainty ellipse.

In addition to the navigation method, the effects of landmark density and landmark detection rate are also studied. The simulated landmark detection rate is implemented by disregarding a specified percentage of the landmark detections. A constant 2-sigma compass uncertainty of $\pm 30°$ is assumed, which is the most conservative value from the Monte Carlo range study with no map information. With 3 different navigation methods, 5 different landmark densities, and 5 different landmark detection rates being considered, a total of 75 different combinations of test conditions are studied. For each combination, 700 simulation trials were conducted to find the success rate as a function of Euclidean distance from starting point to goal. Success rate is defined as the rate at which the vehicle reaches the goal without its 2-sigma uncertainty ellipse major axis exceeding the lost threshold of 100 meters.

Fig. 15 summarizes the performance results from more than 50,000 simulations in this Monte Carlo study. Note that Fig. 15 shows range as Euclidean distance, where Fig. 14 shows Manhattan distance traveled.

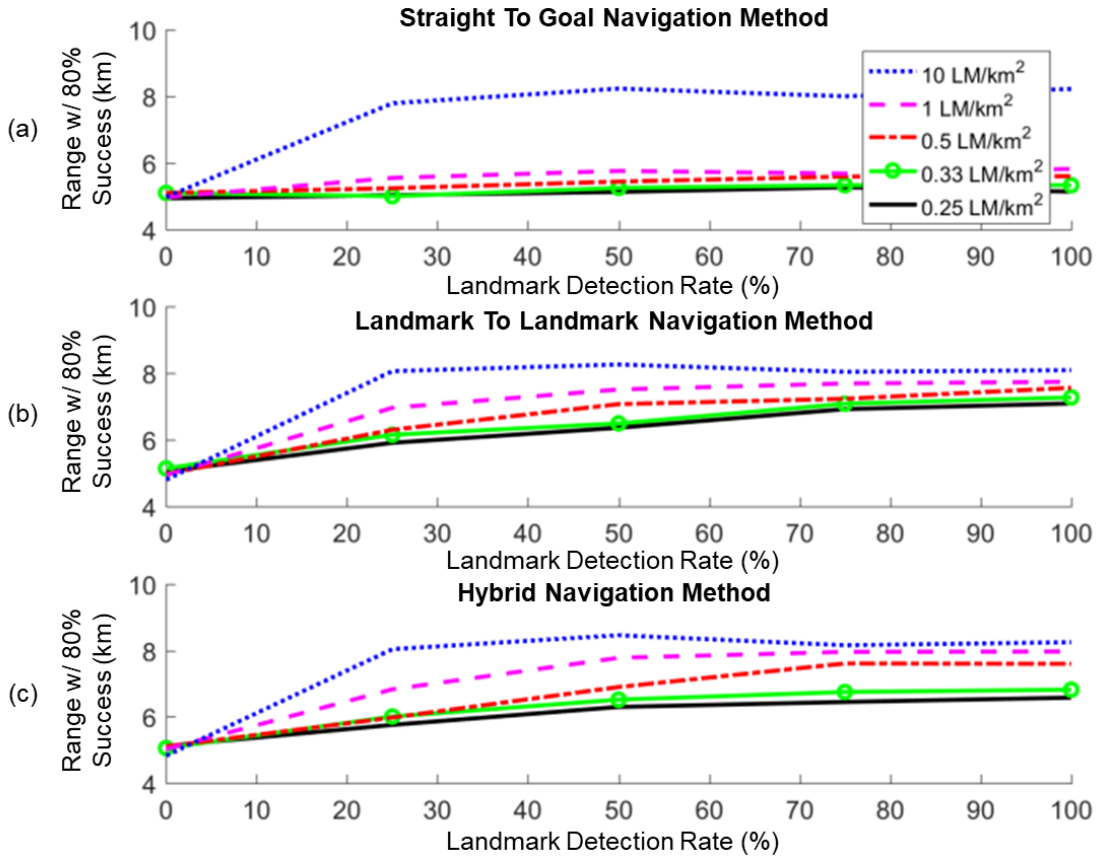

Fig. 15. Monte Carlo landmark detection study results for (a) straight to goal navigation, (b) landmark to landmark navigation, and (c) hybrid navigation; note that this figure shows the range as Euclidean distance traveled by the car, rather than Manhattan distance.

Due to the different navigation methods resulting in different routes to the goal, the Euclidean distance from start point to goal, instead of the total Manhattan distance traveled, is used in Fig. 15 for a fair comparison between navigation methods. The performance defined for this study is an 80 percent success rate of reaching the goal; this success rate is chosen for easy comparison with the experimental results shown in this chapter. For each test condition in this study, the range of the vehicle at a specified success rate gives a good indication of the overall success of the test condition. It is expected that the range and overall success of the navigation method increase as the landmark density and landmark detection rate increase. In general, this trend is seen in the results. For dense landmark maps, a clear upward trend can be seen in the data, with the trend becoming subtler as the density becomes sparser. The trend is also subtler for the straight to goal navigation method, since the vehicle is not actively seeking out landmarks and therefore receives far fewer landmark measurement updates. Overall, Fig. 15 allows for the expected range with a given success rate to be obtained for all combinations of navigation methods, landmark densities, and landmark detection rates.

Fig. 15 also gives insight to the effectiveness of each navigation method. The straight to goal navigation method, where the vehicle ignores the landmarks and attempts to drive straight to the goal, performs noticeably worse compared to the other two methods, which perform similarly in terms of robustness (i.e. the reliability of the vehicle to reach the goal for given test conditions is similar). Excluding the 10 landmarks per sq. km case (in which many landmarks are reached regardless of the navigation method), the range of the straight to goal navigation method for all other test parameters is typically 1 to 2 km less than the corresponding range for the other two navigation methods. However, the navigation methods that actively seek out landmarks (the landmark to landmark and hybrid navigation methods) increase the average distance traveled for the vehicle. The landmark to landmark navigation method causes the vehicle to travel 31 percent further on average compared to the straight to goal navigation method. Similarly, the hybrid navigation method causes the vehicle to travel 15 percent further on average compared to the straight to goal navigation method. Therefore, in general, there is a tradeoff between distance traveled and robustness for the navigation methods that sought out landmarks to update the vehicle's pose estimate.

*Experimental Study for Autonomous Navigation with Limited Information*

To verify the simulation results and understand the maturity of the theory, the proposed estimation and navigation techniques were implemented on a 2007 Chevrolet Tahoe and tested in downtown Ithaca, NY. A picture of the test vehicle is shown in Fig. 18. Due to the current traffic laws in the state of New York, the pose estimation and navigation techniques were implemented and used to guide a human driving the car, directing the driver where to go. In other words, the driver was merely used to act as the inner-loop controller and keep the vehicle on the road.

The odometry measurements for these field tests were obtained from the vehicle's wheel encoder measurements on the CAN bus. A low-cost compass was installed in the vehicle to receive heading measurements. Due to the large amount of electromagnetic interference in the car, the compass' 2-sigma uncertainty was empirically determined to be approximately ±25°. However, this uncertainty is dependent on the location of the vehicle and can improve or worsen based on its location within the city. Finally, a Point Grey Ladybug3 360-degree camera was used to capture images for landmark detection.

*Range Tests without Map Information*

The first set of experiments aimed to verify the results from the Monte Carlo range study with no map information in Fig. 14. For these experiments, the vehicle was driven for a specified amount of time and the final position uncertainty from the pose estimator was recorded. Fig. 16 shows the results from 33 range test experiments with no map information overlaid with the Monte Carlo simulation results.

Results show that the vehicle can travel nearly 10 km with no map information before becoming lost. Given a 2-sigma compass uncertainty of approximately ±25°, the experimental results are expected to fall between the high compass uncertainty and medium compass uncertainty data points. As shown in Fig. 16, this is typically the case, reinforcing the simulation results. However, as the distance traveled increases, the experimental data align more closely with the high compass uncertainty data points. This is likely due to locations where the compass experienced very high magnetic interference, which was seen during calibration tests. As the distance traveled by the vehicle increases, the more likely it is to drive through one of the regions with high magnetic interference, leading to a higher average compass uncertainty.

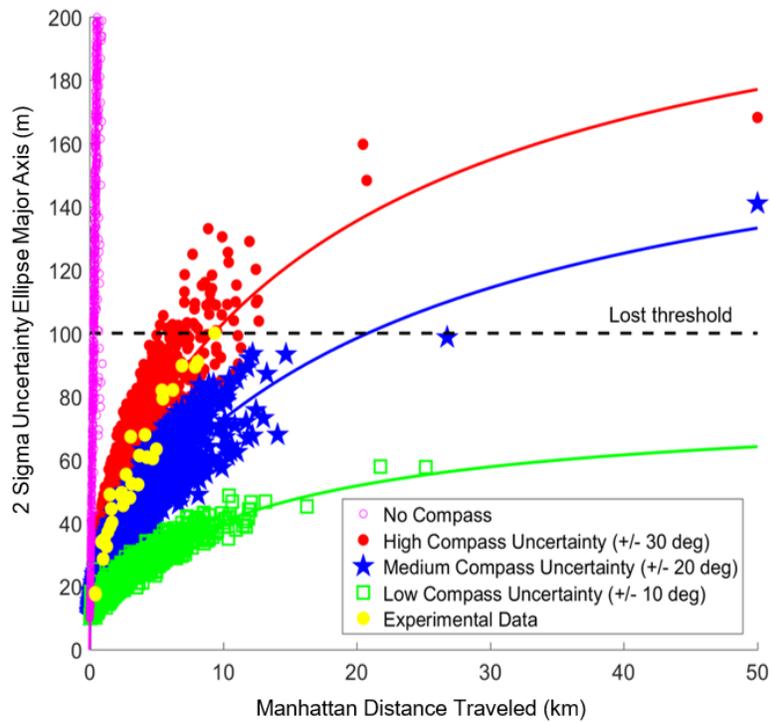

Fig. 16. Experimental results shown overlaid with simulated Monte Carlo results to verify the Manhattan distance an autonomous vehicle can travel without external pose measurement updates before getting lost. Thirty-three field tests were performed for this study.

*Landmark Detection Tests*

To verify the Monte Carlo landmark detection results, an ORB landmark detector was developed and implemented on the vehicle to augment the existing pose estimator. This landmark detector had an average detection rate of approximately 60 percent. Furthermore, approximately 100 images from Ithaca's downtown intersections were obtained to populate the vehicle's landmark database; however, only a few of these images were used during the experiments due to the specific landmark density that was chosen to be tested. Finally, the landmark to landmark navigation method was used in these tests due to its high robustness.

For each test, the vehicle's navigation algorithm guided it to a randomly chosen goal location while relying on the sparse pose estimator for localization. Once the goal was reached, a new random goal was spawned and the vehicle then proceeded to drive to it. This process was repeated until the vehicle became lost and the navigation algorithm broke down or the experiment exceeded 75 minutes. For each goal generated, at most one landmark would be randomly placed within the map. This landmark would then be cleared when the vehicle reached its corresponding goal. Therefore, no more than one goal and one landmark were on the map at a time. This approach of traveling to many subsequent goals, as opposed to one goal, needed to be taken due to the small size of Ithaca's downtown (approximately 1 sq. km). Simulations show that the method of subsequently traveling to many short-range goals produce analogous results compared to traveling to one long-range goal. The supplemental video shows how these experiments were performed.

Ten experiments were conducted with an average landmark density of 0.55 landmarks per sq. km and an average landmark detection rate of 60 percent; the number of tests was chosen based on the time required to run each trial (approximately 1 hour). In 8 of these 10 tests, the vehicle successfully reached a final goal with a Euclidean distance of at least 6.9 km from the starting point. This result is plotted in Fig. 17 along with the Monte Carlo simulation results.

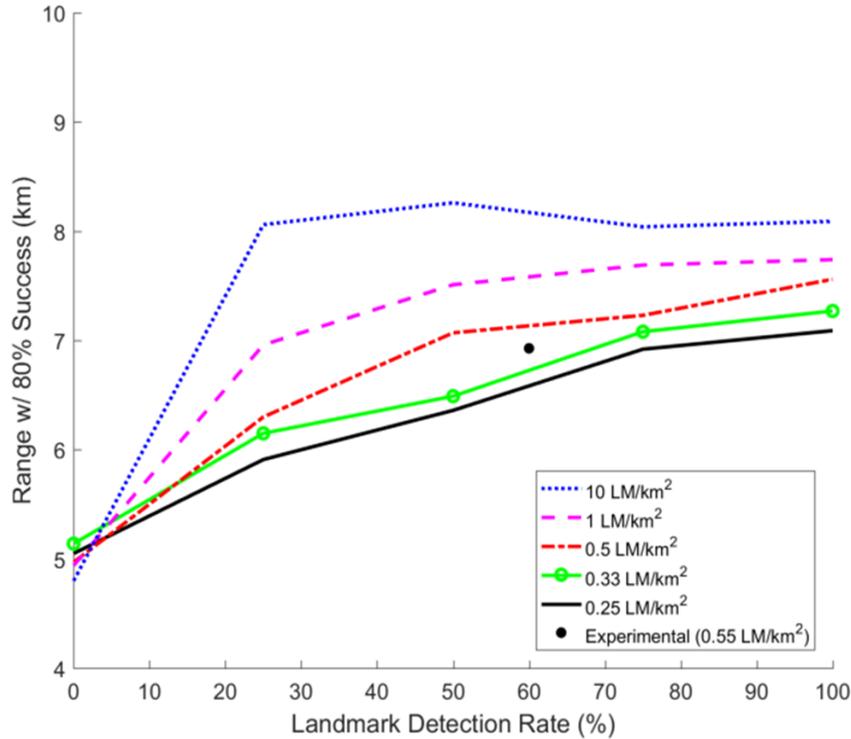

Fig. 17. Experimental result shown with simulated Monte Carlo landmark detection study results. Ten field tests were conducted using the landmark to landmark navigation method to verify the simulation results. Note that this figure shows the range as Euclidean distance traveled by the car, rather than Manhattan distance.

When incorporating a sparse landmark-based map, the vehicle can reliably travel to a much further goal, as compared to the tests without map information. In many of these experiments, the vehicle could drive a Manhattan distance of 20 km (more than twice the distance traveled in the tests with no map information). The experiment showed that the vehicle could drive nearly 10 km between landmark measurement updates without getting lost. Therefore, the experiments demonstrated that, if the landmark density and detection rate were high enough to guarantee a landmark measurement update at least every 10 km traveled by the vehicle, then the vehicle could travel indefinitely without getting lost. This was seen in 3 tests, as the vehicle received numerous landmark measurement updates during its travel; however, these tests were eventually cut short after 75 minutes of driving.

While the average experimental results match the simulation results relatively well, there is some variation. A total of 10 trials is small compared to the number of simulation trials. It is expected that the experimental result would more closely reflect the simulated results with additional tests. Given the large amount of time needed to perform each test, 10 experiments were deemed sufficient to demonstrate the capabilities of the proposed techniques. Additionally, the landmark detection rate varied from test to test. This high uncertainty in the detection rate could also explain why the experimental result is lower than expected in Fig. 17. The detection rate was dependent on many environmental conditions including the prominence of the features at each intersection, the weather, and the traffic. For instance, the vehicle would be less likely to receive a landmark measurement update if the landmark did not have interesting features (e.g. an open field or empty parking lot), or if the weather or traffic was very different from the database images. Fig. 18 shows the test vehicle, as well as a database image and test image of a landmark taken during testing.

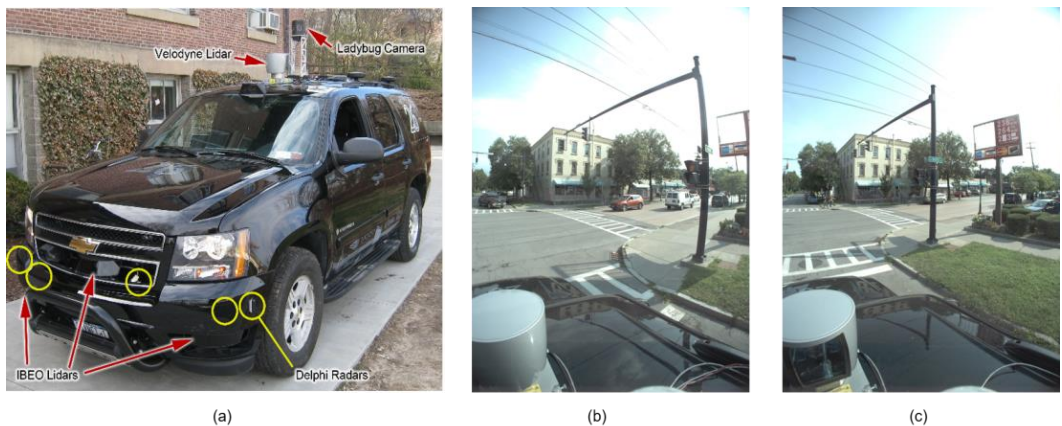

Fig. 18. (a) Test vehicle with notable sensors indicated; note only the Ladybug camera was used in this study, the lidar and radar units were not used. (b) Test image of a landmark from the Ladybug camera during the experiments. (c) Database image of the landmark corresponding to the image in (b). These images produced a match and resulted in a landmark measurement update despite being taken on different days under different conditions. Note the resolution of the original images is 1232 x 1616.

*Conclusions Drawn from Autonomous Navigation Study*

A novel system architecture is presented to address the problem of autonomous driving within an urban environment when reliable GPS measurements and map information is limited. This chapter proposes a pose estimation method that utilizes odometry, compass, and sparse map-based measurements to estimate the pose of the vehicle as it autonomously navigates the roadways with limited map information and GPS measurements. This study also uses a simulator to study key parameters of the navigation and pose estimation algorithms within the proposed system architecture. Monte Carlo studies using this simulator provide evidence to resolve key issues concerning navigating without GPS or detailed map information. These studies show the distance a vehicle can travel with no GPS or map information, as well as the relationship between the range of the vehicle, navigation method, landmark density, and landmark detection rate. Experimental results verify the simulation results within a small amount of deviation, as they produce a minimum range of 6.9 km for the given success rate, navigation method, landmark detection rate, and landmark density.

CHAPTER 3

REAL-TIME STATIC SCENE ESTIMATION FOR ROADWAY INTERSECTIONS

*Introduction to Static Scene Estimation*

Acquiring a high-definition (HD) map that is sufficiently detailed and precise to enable autonomous driving capabilities remains a major hurdle to overcome when developing an autonomous vehicle. Many of the front-running companies in this field, such as Waymo, Uber, and Ford, dedicate entire teams to produce and maintain these HD environmental maps, which include painted road lines and markings, barriers, traffic lights, traffic signs, buildings, and even potholes and other road quality data. The primary method of obtaining such a map involves deploying sensory vehicles to acquire 360-degree lidar and/or camera data of a certain environment multiple times [55], [56]. It is only after this data is collected, followed by heavy post-processing to create the final HD map, that a vehicle can drive autonomously using the precise map in the given environment. While recent developments utilizing a pose-based GraphSLAM estimator provide even more precise mapping, these approaches remain highly time, cost, and labor intensive [57]. An alternative method of obtaining an HD map of the environment involves the use of unmanned aerial vehicles, equipped with lidar and camera systems, as a novel platform for photogrammetry [58]. However, this method also requires time-consuming data collection with heavy data post-processing. Due to the extreme complexity and laborious nature of the HD mapping task, automakers such as Volkswagen, BMW, and General Motors have relied on third-party services, such as HERE and MobilEye, to provide these HD maps [59].

Another difficulty associated with HD mapping is the brittle nature of the maps. An HD map is highly detailed and accurate for a given area at the time it is originally generated. However, there is no guarantee that the map will continue to be sufficiently detailed and accurate in the future. Therefore, once acquired, HD maps must undergo continual maintenance due to their high resolution and sensitivity to environmental changes, such as resurfacing roads, repainting lane markings, lane or road closings, and other construction related tasks. This continual need for maintenance further adds to the time, money, and resource costs associated with using HD maps to enable autonomous driving capabilities.

While HD maps could be stored for local driving and *a priori* routes, it is infeasible to store a high-fidelity map onboard the vehicle for all locations, particularly when considering environments such as cities that experience rapid change and/or development. Data connections could be relied upon to provide the HD map of the environment; however, these connections can be weak, spotty, or non-existent in urban areas. Furthermore, security of autonomous vehicles is a major concern, as the vehicle's map can be spoofed and/or altered out of malicious intent [60]. Despite there being several drawbacks to HD maps for navigation, there is no alternative for intelligent mobility solutions in the current state of technology [61]. To make autonomous vehicles practical for a wide range of environments, a more scalable method of providing environmental map information must be developed. This study investigates an architecture that utilizes minimal *a priori* information about the environment, and instead relies on real-time sources of information from sensors to enable robust navigation in urban environments.

There have been various recent research efforts that aim to allow for autonomous driving capabilities without the use of *a priori* HD mapping. Online *Simultaneous Localization and Mapping* (SLAM) can be applied to autonomous driving to alleviate the need for high-fidelity maps [62], [63]. For example, FAB-MAP, a topological SLAM technique, allows for appearance-based navigation [64]. Similarly, SeqSLAM is another SLAM technique that allows for appearance-based navigation, but also aims to address the difficulty associated with changing environmental conditions [65]. Although both methods do not rely on an *a priori* HD map, they operate purely in appearance space and make no attempt to localize the vehicle in metric coordinates, either global or map-based; in other words, these methods act as scene-matching algorithms. Without the ability to localize the vehicle in metric coordinates, it is impossible to localize the vehicle when it is in between two matched scenes. Therefore, a sufficiently dense map of images is required to adequately localize the vehicle. As the size of the environment grows, scalability becomes a major concern due to sharply rising computational requirements and rapidly diminishing performance [66]. To address the issue of localizing the vehicle in metric coordinates, GPS-fused SLAM techniques have been proposed [67], [68]. However, these methods still struggle in terms of scalability. Furthermore, they rely on consistent, high precision GPS measurements, which are often unreliable in areas such as dense urban environments. Alternative methods [69], [70], [71], which take a perception approach, demonstrate the ability to automatically map the 3D position of traffic lights, enable real-time autonomous off-road navigation using semantic mapping, and allow for real-time joint semantic reasoning for street scene understanding. These techniques, however, remain limited in terms of real-time mapping, as they do not provide the capability to infer the full roadway intersection scene required for navigation.

The goal of this work is to develop a novel real-time scene estimation framework that is sufficiently robust to allow for autonomous navigation without heavy reliance on a detailed *a priori* map – in other words, without requiring detailed roadway information, such as road lines and markings, barriers, traffic lights, traffic signs, and other various roadway data, all to be known in advance. This study is a continuation of the work described in [72], which presents a localization and navigation strategy, as well as an overall system architecture, to address the issue of autonomous driving with minimal information (i.e. no GPS measurements or detailed *a priori* map information). The method proposed in this chapter utilizes computer vision techniques to detect environmental cues which, when reasoned together in a probabilistic framework, the static scene around the vehicle can be inferred in real time. Due to the unique navigation challenges presented by urban roadway intersections, as well as their importance to overall navigation, the experimental validation for this work focuses specifically on 3-way and 4-way intersection scenes. While the primary focus of this chapter is on navigation without the use of an *a priori* HD map of the environment, this work may also be used to supplement current navigation systems that use *a priori* maps and/or SLAM techniques. This supplemental technique could provide vital support to autonomous vehicles for cases in which the navigation system experiences difficulties, such as during a GPS blackout in a dense urban environment, or has its security threatened, such as during an attack by a cybersecurity adversary. This robustness to infrequent, yet crucial events is critical for long-term navigation solutions.

*System Architecture for Intersection Inference*

A system architecture for autonomous navigation without GPS measurements or high-definition *a priori* map information is shown in Fig. 19. This system diagram contains familiar elements in autonomous driving, such as steering and speed controllers, an object tracker, and a path generator. In [72], the intersection navigation algorithm and the lightweight sparse pose estimator – which only relies on odometry, compass, and sparse landmark measurements – are described in detail to address the challenges associated with the lack of detailed map information and GPS measurements. However, this previous work assumes that a static estimate of the scene is available. This chapter expounds upon the architecture by eliminating this assumption and further developing the static roadway scene estimator component.

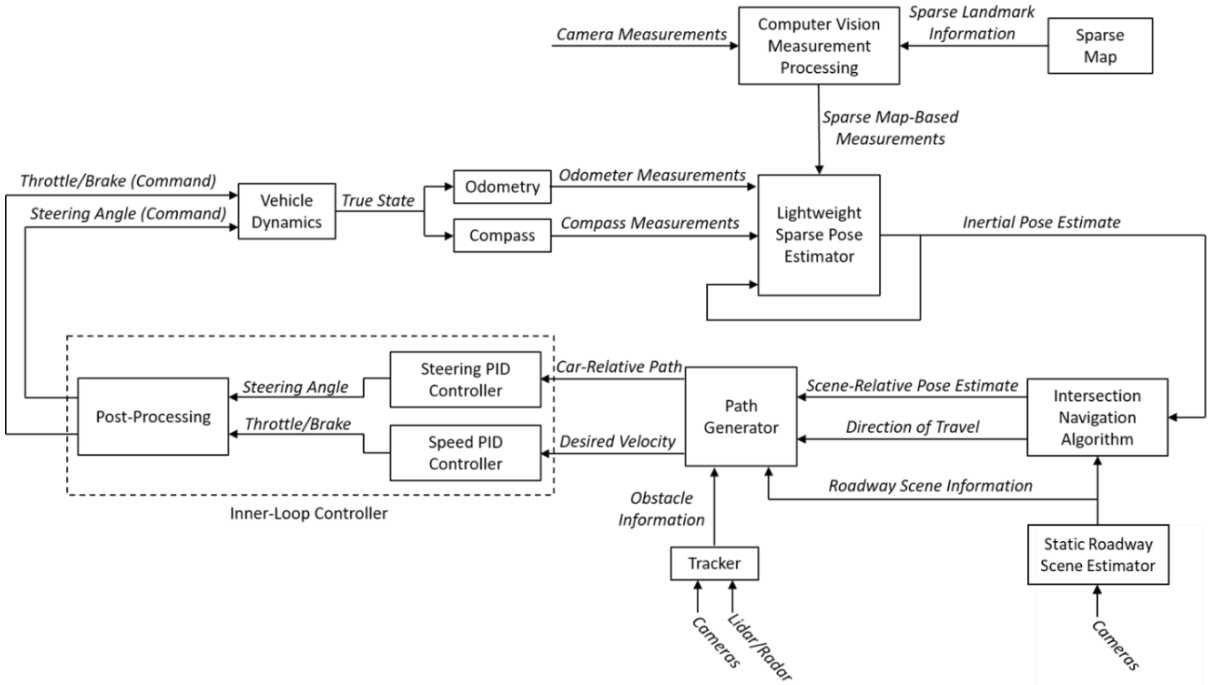

Fig. 19. System architecture for autonomous navigation with minimal information (i.e. no GPS measurements or detailed *a priori* map information).

To enable autonomous navigation without HD *a priori* map information, the vehicle must be able to detect and classify key roadway features in real-time in order to infer key map attributes. Such key features required for navigation include lane line location and type, intersecting roads, turn options at intersections, traffic lights, stop signs, and other navigation-critical street signs, such as one-way and do not enter signs. Fig. 20 shows the detailed pipeline for the proposed real-time static roadway scene estimator, seen in Fig. 19. A raw image from the vehicle's front-facing camera is the input to the pipeline. This image is then passed to several computer vision components which extract environmental cues to infer an understanding of the static scene to a sufficiently detailed level for autonomous navigation. Table 5 summarizes the environmental cues used for this study. The output of the static scene estimator are lane line locations and types, an estimated car-relative intersection location, and probabilities for key intersection features. These features include the intersection itself, as well as its turning options (i.e. left turn, right turn, or continue straight). A marginal probability is computed for each feature in real-time.

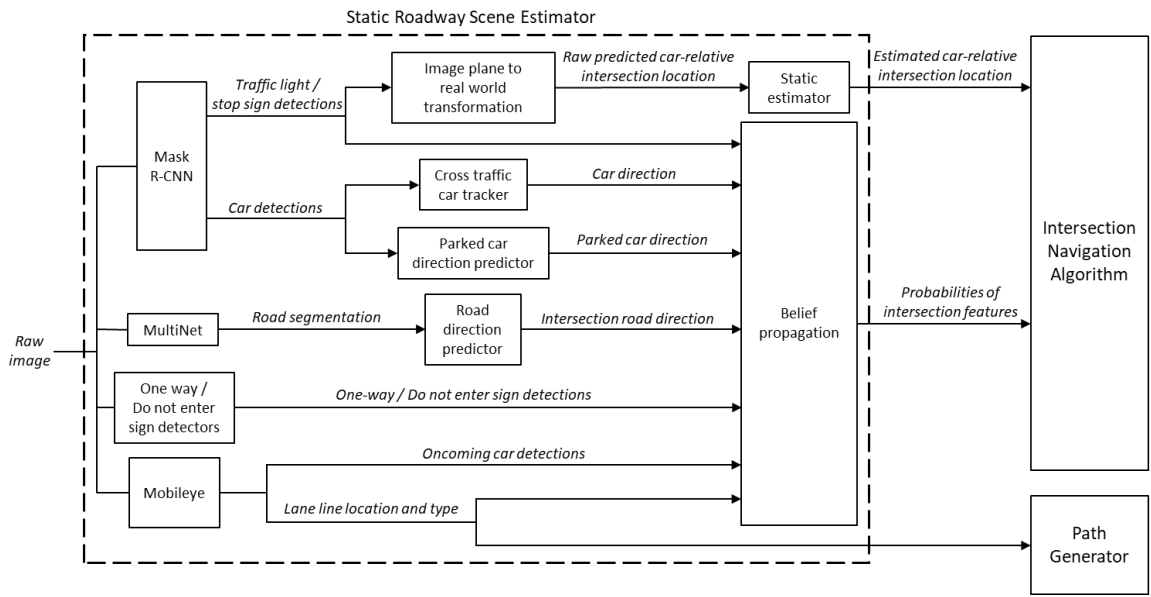

Fig. 20. Pipeline for the real-time static roadway scene estimator.

Table 5. Summary of environmental cues used for static roadway scene estimation.

| Environmental Cues | Detection Method | Contribution to overall scene understanding |
|---|---|---|
| Lane lines | Mobileye 560 | Lane line type, shape, and location relative to vehicle |
| Traffic light | Detectron | Presence of intersection, and approximate intersection location relative to vehicle |
| Stop sign | Detectron | Presence of intersection, and approximate intersection location relative to vehicle |
| Road surface | MultiNet | Possible directions of travel at intersection |
| Cross traffic vehicles | Detectron | Possible turning options at intersection |
| Parked vehicles along crossroad | Detectron | Possible turning options at intersection |
| Oncoming/outgoing vehicles | Mobileye 560 | Possible direction of travel at intersection |
| One-way signs | Sliding window object detection | Unviable turning options at intersection |
| Do not enter sign | Sliding window object detection | Unviable direction of travel at intersection |

*Intersection Estimator*

Strong environmental cues are used to reliably detect and locate an intersection relative to the vehicle. These strong cues include stop sign and traffic light detections, since these are most conspicuous for detecting intersections. Furthermore, due to stop sign and traffic lights' close proximity to the intersection itself, their estimated car-relative location can be used as an approximate location for the intersection also. To allow for car-relative location estimation, the stop signs and traffic lights must be detected with a high level of precision. To attain this high level of precision, the proposed method uses Detectron – an open-source, state-of-the-art object detection platform developed by Facebook AI Research, which implements advanced object detection algorithms, such as Mask R-CNN [73], [74]. Using Detectron's implementation of Mask R-CNN trained on the COCO dataset, common roadway objects, such as traffic lights, stop signs, and cars, can be accurately detected and classified within the proposed architecture [75]. The output of Detectron includes a mask of pixels covering the detected object in the image, a label for the object, and a score between 0 and 1. For this study, only detections with a score greater than 0.9 were considered. Using this method, a detector was developed for this work which outputs two binary variables, shown in (7), indicating the presence ($z_{TL} = 1$ or $z_{SS} = 1$) or absence ($z_{TL} = 0$ or $z_{SS} = 0$) of a stop sign or traffic light. Using the 0.9 score threshold and comparing the detections found during testing with truth data, detection precisions of 0.97 and 0.95 were found for stop signs and traffic lights, respectively. These detection precisions, as well as the detector's outputs, determine the messages described later in the belief propagation section of this chapter. Fig. 21 shows example stop sign and traffic light detections found during testing.

$$\tilde{z}_{TL/SS} = \begin{bmatrix} z_{TL} \\ z_{SS} \end{bmatrix} \qquad (7)$$

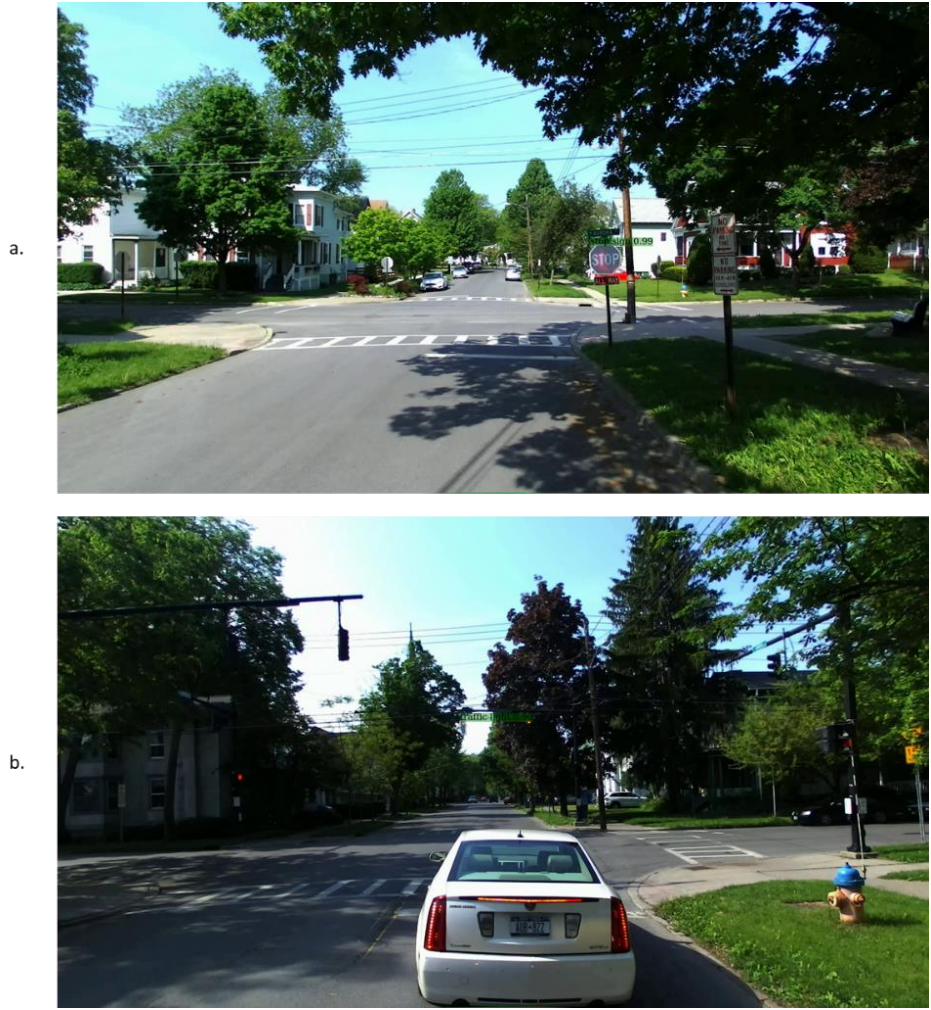

Fig. 21. Example detections using Detectron for (a) stop sign, with a score of 0.99, and (b) traffic light, with a score of 0.90.

To obtain a car-relative intersection location estimate, the segmented shapes for both traffic lights and stop signs are transformed from the image plane to a car-relative position. These car-relative locations are approximated by using a relationship between the segmented shape's height in the image plane and its relative position to the vehicle. This relationship was determined through a data-driven approach, in which traffic light and stop sign segmentations were computed at known distances from the object in thousands of images at various intersections under diverse conditions. This data, as well as the fitted curve that describes the relationship between segment height and distance from detected object, is shown in Fig. 22. The predicted distance to stop sign had a standard deviation of 1.7 m and the predicted distance to traffic light had a standard deviation of 5.9 m. For this study, the predicted distance to a stop sign was considered an approximate distance to the start of the intersection, and the predicted distance to a traffic light was considered an approximate distance to the end of the intersection. For measurements that returned multiple detections, the mean predicted distance was used. To generalize this to account for a wider variety of intersections, beyond the intersections considered in this study, additional crossroad cues would need to be incorporated to determine the intersection location.

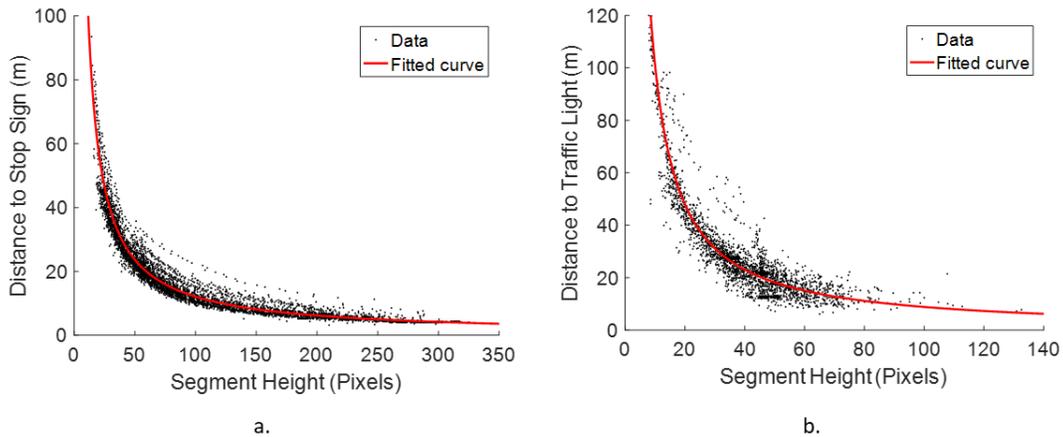

a.  b.

Fig. 22. Data, as well as fitted curve, relating segmented height to vehicle's distance to (a) stop sign and (b) traffic light for various intersections under diverse conditions. The predicted distance for stop sign had a standard deviation of 1.7 m and the predicted distance for traffic light had a standard deviation of 5.9 m.

The raw predicted car-relative intersection locations then pass into a 1-D static estimator, which fuses measurements from the vehicle's odometry, to provide a car-relative intersection location estimate in real-time. The relevant equations for this static estimator are shown below.

$$\bar{x}_k = \hat{x}_{k-1} - dt \tag{8}$$
$$\bar{\sigma}_k^2 = \sigma_{k-1}^2 + q \tag{9}$$

$$K_k = \frac{\bar{\sigma}_k^2}{\bar{\sigma}_k^2 + r} \tag{10}$$

$$\hat{x}_k = \bar{x}_k + K_k(z_k - \bar{x}_k) \tag{11}$$

$$\sigma_k^2 = \bar{\sigma}_k^2 - K_k \bar{\sigma}_k^2 \tag{12}$$

In (8), $\hat{x}_{k-1}$ is the 1-D car-relative location estimate of the intersection from the rear axle of the vehicle at time step $k-1$, $dt$ is the distance traveled by the vehicle between time steps (obtained from the vehicle's odometry), and $\bar{x}_k$ is the predicted location estimate at time step $k$. In (9), $\sigma_{k-1}^2$ is the location estimate error variance at time step $k-1$, $\bar{\sigma}_k^2$ is the predicted location error variance at time step $k$, and $q$ is the odometry noise variance. The location estimate and error variance are updated in (11) and (12) after receiving a measurement from the detected stop sign or traffic light, $z_k$. $K_k$ is a Kalman gain from (10), and $r$ is the measurement noise variance.

For intersections that lack the conspicuous features of traffic lights and stop signs, road surface detections (as described in the next subsection) are relied upon to establish the presence of an intersection. Then once the presence of an intersection is established, the remaining cues are activated to infer the key features needed to navigate the intersection. Prior to detecting an intersection, the remaining cues are inactive to mitigate the effects of false positives.

*Road Segmentation and Direction Prediction*

As one of the cues to determine the possible directions of travel at an intersection, road segmentation is used to detect the road surface in an image. Then, based on the shape of the road segmentation, possible directions of travel are inferred. MultiNet is used to perform real-time semantic reasoning of the roadway scene [71]. A pre-trained model using the Kitti dataset was fine-tuned to be used in this study by adjusting threshold parameters and image characteristics.

The output of MultiNet is a segmented image of all pixels where road surface is detected. Based on the location of these pixels in the image, possible directions of travel at an intersection can be predicted. This prediction is performed by examining three regions of interest in the image where road surface is expected to be for an intersection: straight, left, and right. Fig. 23 shows the output of MultiNet, as well as the raw RGB image, for two example cases; the regions of interest are highlighted in all images. A specific road direction is predicted to exist if more than 20 percent of the pixels in the corresponding region of interest were detected to be road. For instance, in Fig. 23a, more than 20 percent of the pixels in the left and right regions are predicted to be road, while less than 20 percent of the pixels in the center region are predicted to be road. Therefore, for the case shown in Fig. 23a, a road direction of left and right were predicted, while straight was not. This prediction does not attempt to measure any characteristics of the detected road, but strictly predicts whether there exists road in a given direction that could be traveled upon. This condition is described in (13), where $\text{ROI}(p_{ij} = 1)$ represents the count of pixels where road surface is detected in the region of interest, and ROI represents the total pixels in the given region of interest. The output of this detector, shown in (14), are three binary variables representing the detected road directions at the intersection – right, straight, and left. From comparing this detector's output during testing to truth data, a detection precision of 0.77 was found. This comparison was made by checking truth data to see whether a positive detection of a road direction did, in fact, correctly reflect the true environment around the vehicle. Lighting conditions and nearby driveways and parking lots were the primary factors for much of this detector's difficulty. The messages described later in the belief propagation section of this chapter were determined using this detector's output and its detection precision.

$$\frac{\text{ROI}(p_{ij} = 1)}{\text{ROI}} > 0.2 \tag{13}$$

$$\bar{z}_{\text{road}} = \begin{bmatrix} z_{\text{road}_R} \\ z_{\text{road}_S} \\ z_{\text{road}_L} \end{bmatrix} \quad (14)$$

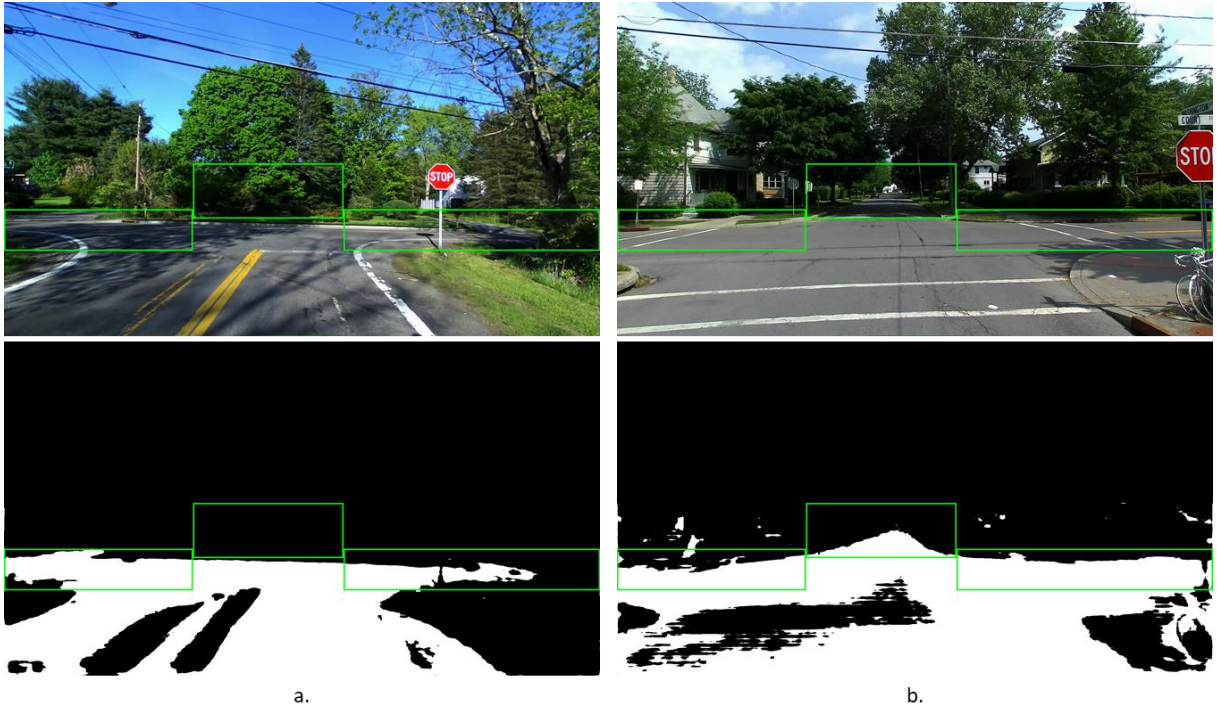

Fig. 23. MultiNet output, as well as raw RGB images, for a (a) 3-way intersection and (b) 4-way intersection. In (a), left and right turning options were detected, but not an option to continue straight. In (b), an option to turn left, right, and continue straight were detected.

*Cross Traffic Car Detection and Tracking*

Detecting and tracking the movement of cars at an intersection is a cue that provides information regarding the manner in which the intersection can be navigated. For instance, if an autonomous car is approaching an intersection and detects other cars travelling from left to right across its field of view, then a reasonable inference to make is that the autonomous car has the option to turn right at that intersection. With this rationale, segmented car shapes from Detectron are passed into a cross traffic car tracker to determine the possible directions of travel at an intersection. Fig. 24 shows car detections from consecutive time steps that were obtained during testing. While any object tracker would work [76], a stand-alone tracker was developed by comparing consecutive segmented frames and inferring the direction of the segmented cars by their movement over the consecutive frames. The correspondence between two segmented shapes in two consecutive frames was determined by comparing the shapes' sizes and locations in the images. If the shapes' sizes differed by less than 3 percent and its locations differed by less than 5 percent of the total image width, then the shapes were assumed to be from the same object. To mitigate false tracks, the tracker only considered car segments with a length-to-height ratio greater than 2 and a score above 0.95. The length-to-height ratio was used to find only the segmented car shapes that represent the side profile of a car, which generally indicates that the car belongs to the cross-traffic. The score threshold was used to filter out partially occluded cars, which generally provided lower scores. To further mitigate false tracks, the tracker only considers tracks that indicate steady, horizontal motion across the image between consecutive frames (i.e. the expected motion of cars belonging to the cross-traffic). This filters out parked cars in parking lots, driveways, and on the street. The output of this cross traffic tracker, $\tilde{z}_{ct}$, shown in (15), are two binary variables representing the presence of a right-moving vehicle and a left-moving vehicle for any two consecutive frames. These binary variables were 0 for no detection of cross traffic, and 1 when a segmented car shape met all the previously described conditions for a given direction. From comparing the tracker output during testing to truth data, which involved 80 tests with up to 100 detections in a given test, a detection precision of 0.86 was found. This detection precision, as well as the detector's outputs, determine the messages described later in the belief propagation section of this chapter.

$$\tilde{z}_{ct} = \begin{bmatrix} z_{ct_R} \\ z_{ct_L} \end{bmatrix} \tag{15}$$

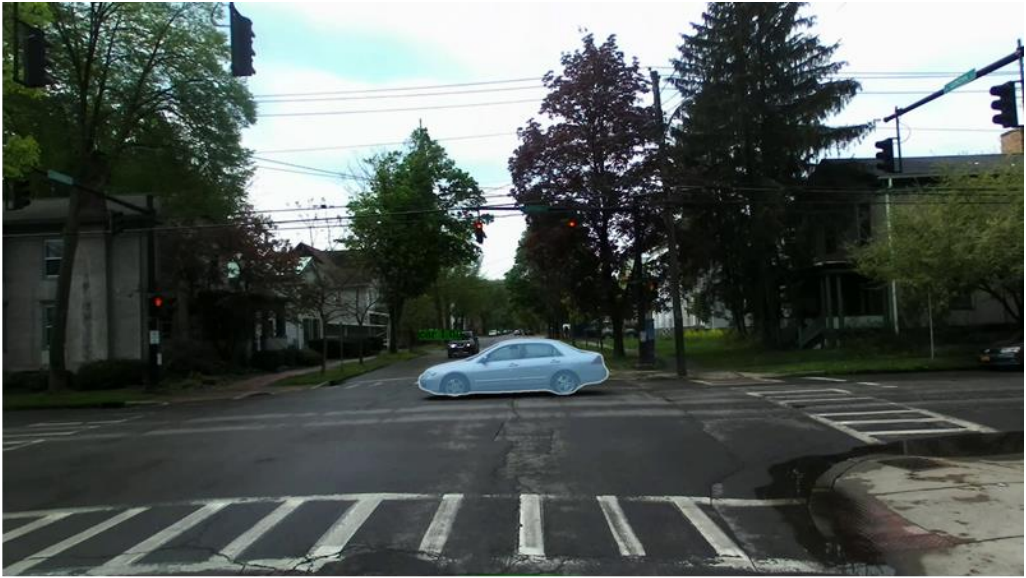

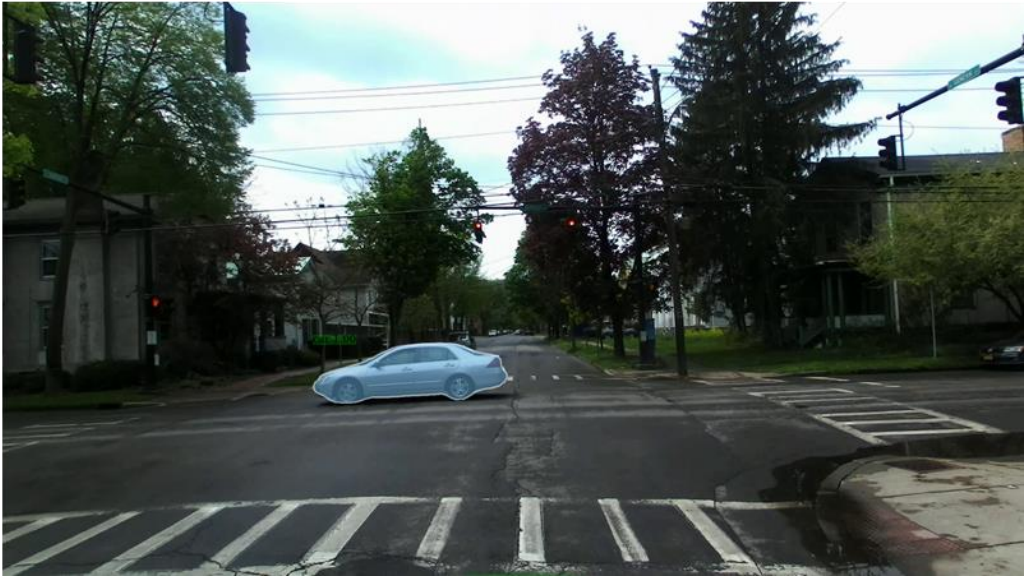

Fig. 24. Detectron cross traffic detections from consecutive time steps, (a) and (b). These detections resulted in a positive left cross traffic measurement due to the segmented vehicle's predicted direction of travel, based on its size, geometry, and horizontal motion.

*Parked Car Detection and Orientation Prediction*

Similar to detecting the direction of cross traffic at an intersection, detecting the orientation of parked cars along an intersecting road can give insight regarding the possible directions of travel at an intersection. For instance, the parked cars along the intersecting road seen in Fig. 25 indicate that turning left is a viable option at that intersection. The same size, geometry, and score constraints that were used for the cross traffic detector were also used for this detector to filter oncoming, occluded, and other undesired vehicle detections. To filter moving cars, this detector only considers detections of cars that did not show evident horizontal or vertical motion across the image between consecutive detections. Once the segmented car shapes which met these conditions are obtained, their orientation is predicted by evaluating their shape. Due to the standard aerodynamic shape of vehicles, the nose of the car can predictably be differentiated from the tail of the car by comparing their heights. The end of the car with the lower height is assumed to be the nose of the car. Shown in (16), the output of this parked car detector, $\tilde{z}_{pc}$, are two binary variables representing the presence of a right-facing parked car and a left-facing parked car on the crossroad. These binary variables are 0 for a given direction when no parked cars are detected in that direction, and 1 for a given direction when a segmented car shape meets all the previously described constraints (size, geometry, score, and motion) and predicts the corresponding orientation based on its geometry. Comparing this detector's output during testing to truth data, which involved 80 tests at various intersections, a detection precision of 0.83 was found. The messages described later in the belief propagation section of this chapter were determined using this detector's output and its detection precision.

$$\tilde{z}_{pc} = \begin{bmatrix} z_{pc_R} \\ z_{pc_L} \end{bmatrix} \qquad (16)$$

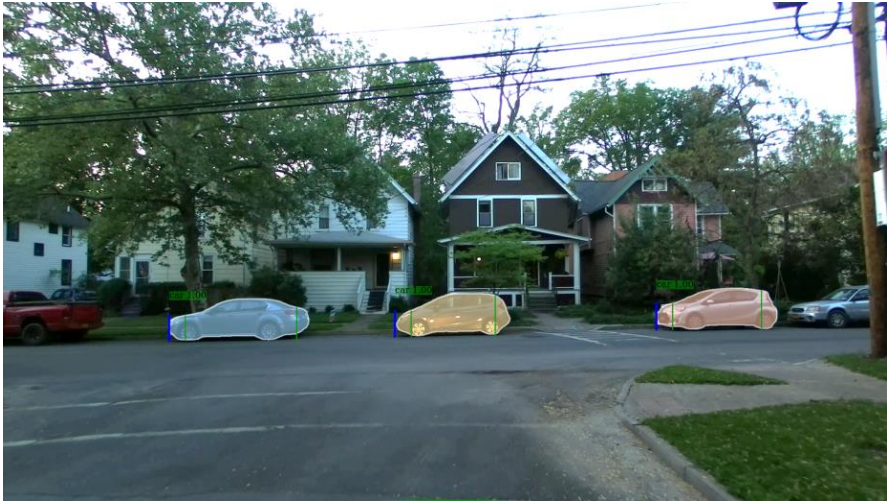

Fig. 25. Detections of parked cars using Detectron, and predictions of their orientation. These detections resulted in a positive left turn measurement due to the predicted orientation of the segmented vehicles, based on their geometry. In the figure, the green lines on the segmented shapes indicate the measured heights of the right and left end of the vehicle, and the blue line at the front of the segmented shape indicates the resulting prediction for the nose of the vehicle.

*Lane Line and Oncoming/Outgoing Car Detection*

Additional environmental cues that can be used to detect possible directions of travel at intersections are lane lines and oncoming or outgoing cars. For this study, Mobileye 560 was used to provide lane line and obstacle data in real time [77]. The lane line data is given as a second order polynomial function relative to the ego vehicle, and the obstacle data includes type (e.g. vehicle, pedestrian, cyclist), relative longitude and lateral position, and relative longitude velocity. The lane line information is primarily needed for the low-level path generator (see Fig. 19 and 20) but can also be used to detect the option to continue straight at intersections that have easily visible lane lines from across the intersection. Furthermore, detecting oncoming or outgoing traffic provides cues that there is a possibility to travel straight at an intersection. Mobileye detections are used just to detect oncoming or outgoing traffic, as opposed to detecting cross traffic, due to the device's narrow field of view. An example of this detection is shown in

Fig. 26. These Mobileye detections are used to infer a binary variable representing the option to continue straight at an intersection. If lane lines are detected to continue straight past an intersection, the binary variable $z_{lanes} = 1$ is passed on through the pipeline. If an oncoming or outgoing vehicle is detected past an intersection, the binary variable $z_{on/out} = 1$ is passed on through the pipeline. According to Mobileye documentation, all detections have a precision of 0.99. However, since this study is using its detections to predict road direction at an intersection, this precision is lower due to false measurements caused by cars parked in driveways and parking lots. From comparing the predicted road direction from these measurements to truth data during testing (80 tests at various intersections), a detection precision of 0.96 was found. This detection precision, as well as the detection outputs, determine the messages described later in the belief propagation section of this chapter.

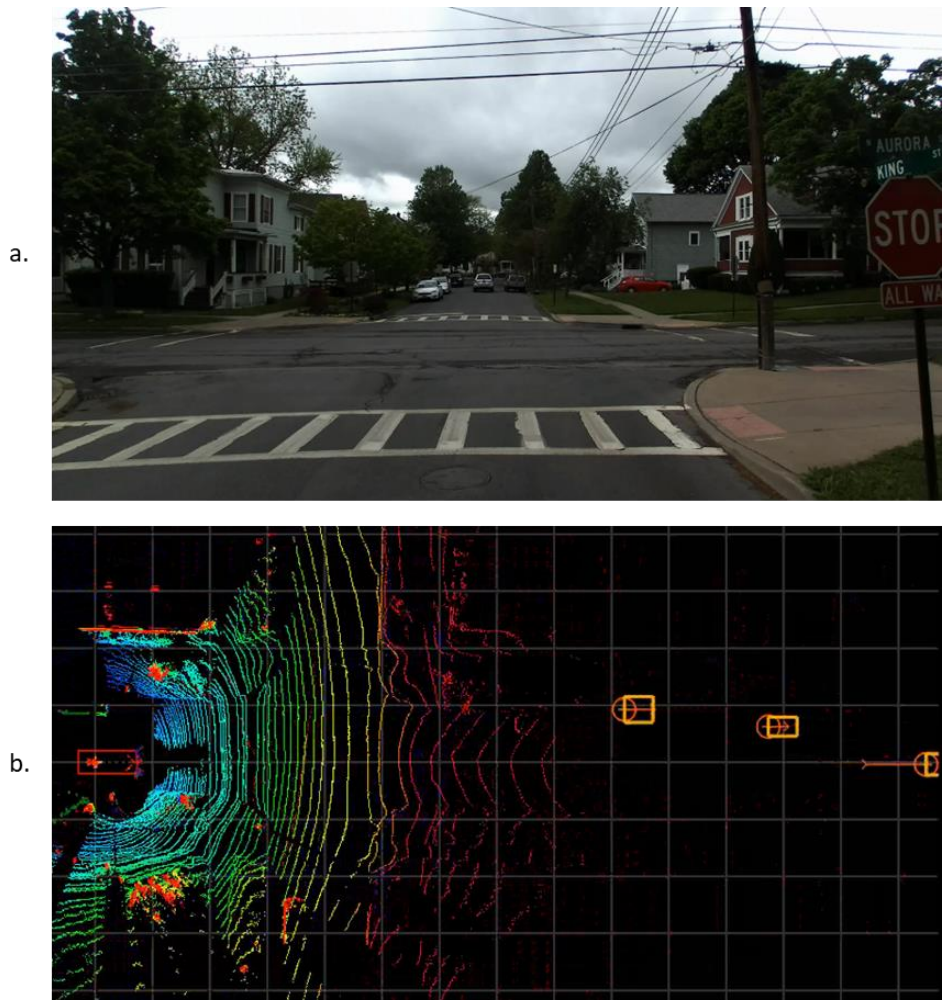

Fig. 26. Mobileye detections of oncoming and outgoing vehicles. (a) Raw RGB image of the intersection scene. (b) Sensor output from the test vehicle. Mobileye detections of the two white cars parked along the side of the road, as well as the outgoing silver car, are shown by the yellow boxes.

*One-Way and Do Not Enter Sign Detection*

Not all directions of travel that are predicted using the previously described environmental cues are always possible options for travel. For instance, a car can detect a road to the left at an intersection and therefore predict that the intersection contains an option to turn left. However, if there exists a one-way sign in the opposite direction of this road, then this left-hand turn option is not a viable possibility. Therefore, a one-way sign detector and do not enter sign detector were developed for this work to ensure robust navigation capabilities. These detectors use the sliding window detection technique by scanning an image using a pyramid representation and extracting ORB features [78]. These extracted features from the sliding window are then matched with the features from a sample image of the sign using the ratio test [79]. The threshold for the ratio test was tuned to ensure that the detection precision was above 0.95. Although tuning to this specification resulted in a lower recall, this was deemed acceptable to maintain a high precision. To increase computation speed, the top and bottom areas of the image are cropped out, leaving only the area where one-way and do not enter signs are typically seen. The output of the do not enter sign detector is a binary variable, $\tilde{z}_{DNE}$, that is 1 when a do not enter sign is detected and 0 when one is not. The output of the one-way sign detector are binary variables representing the detection of left or right one-way signs, shown below in (17). These outputs, as well as the detection precision, determine the messages described later in the belief propagation section of this chapter.

$$\tilde{z}_{ow} = \begin{bmatrix} z_{ow_R} \\ z_{ow_L} \end{bmatrix} \qquad (17)$$

*Belief Propagation*

Once the environmental cues are obtained from the detection techniques described in the previous sections, the presence of key intersection features for navigation can be probabilistically determined. For this study, these features included the presence of an intersection itself, as well as its possible directions of travel (i.e. left, right, or straight). Belief propagation is used to compute the probabilities associated with each intersection feature. A graph displaying the manner in which the observed environmental cues, $z$, relate to the latent intersection features, $f$, is shown in Fig. 27. In the graph, the nodes along the top represent the latent intersection features, and the nodes along the bottom represent the observed environmental cues. Black edges connect cues and features that are directly related, while red edges connect cues and features that are mutually exclusive. The factor graph is segmented into three independent sub-graphs and the corresponding probabilities are computed separately. The marginal probability of each intersection feature is initialized to 0.5. At each time step, a message, $\mu_i$, is passed from each observed cue to its corresponding intersection feature. The values of the messages correspond to the detection probabilities described in the previous sections, whether the observed cue is detected, and whether the cue and feature are directly related or mutually exclusive. For the parked car, oncoming/outgoing car, and cross traffic cues, each detection resulted in a sent message. Therefore, if multiple left-facing parked cars are detected, multiple messages would be sent to the left turn node, resulting in a higher left turn probability. If an environmental cue is not detected, a neutral message ($\mu_i = 0.5$) is sent. Until the presence of an intersection is established by the detections of a traffic light, stop sign, or right/left road surface, the messages from the remaining cues are neutral. The presence of an intersection is established once the probability associated with the intersection feature or the right/left turn feature exceeds 0.9. At this point, the remaining cues are activated to detect the remaining intersection features. Finally, if an environmental cue is not detected after the vehicle comes within 20 m of the intersection, then a very weak dissenting message ($\mu_i = 0.49$) is sent to the node for the corresponding direction of travel. The marginal probability of a given feature is computed at each time step by the product of all incoming messages to the given feature node, as shown in the following equation:

$$p(f_j) \propto \prod_{i \in \mathcal{N}(j)} \mu_i(f_j) \qquad (18)$$

In the above equation, $p(f_j)$ is the marginal probability of feature $f_j$, $\mu_i(f_j)$ is the message from an observed cue to feature node $f_j$, and $\mathcal{N}(j)$ represents the set of all cues that are linked to feature node $f_i$.

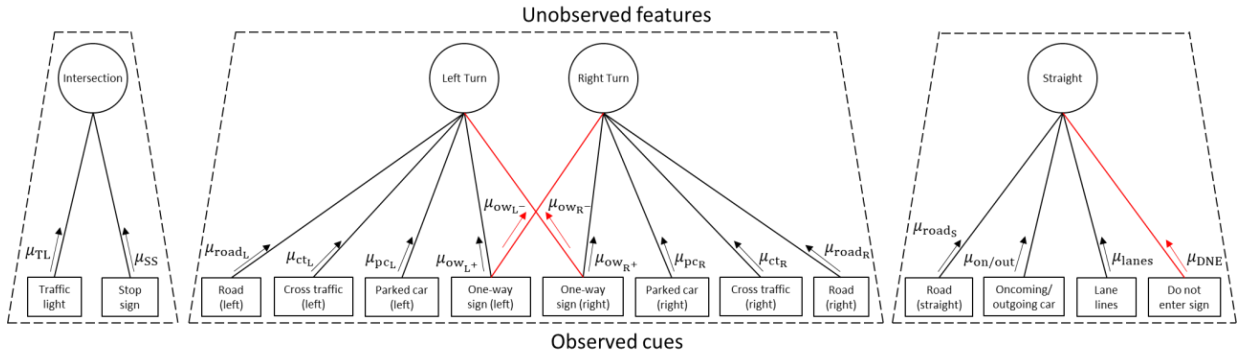

Fig. 27. Graph describing the relationship between observed environmental cues along bottom, $z$, and latent intersection features along top, $f$.

*Experimental Study of Intersection Scene Estimation*

An experimental study to test the performance of the proposed technique was conducted with Cornell University's autonomous test vehicle in downtown Ithaca, NY. A picture of the test vehicle is shown in Fig. 28. For this study, 80 tests were performed in which the test vehicle approached an intersection while gathering image and odometry data and running the proposed algorithm to estimate the static roadway scene. The 80 tests included 10 tests from 8 different intersections. These field tests included both 3-way and 4-way intersections, as well as various roadway features, such as traffic lights, stop signs, and one-way signs. A summary of the 8 test intersections is shown below in Table 6. The 10 tests performed at each intersection were carried out under diverse environmental and vehicle conditions, such as lighting, weather, traffic, and vehicle speed. A front-facing ZED stereo camera was used to obtain the raw images of the roadway scenes. Odometry measurements for these field tests were obtained from the vehicle's wheel encoder measurements on its CAN bus. Computations for the detection and segmentation algorithms were performed on TITAN X and 1080 Ti graphics cards.

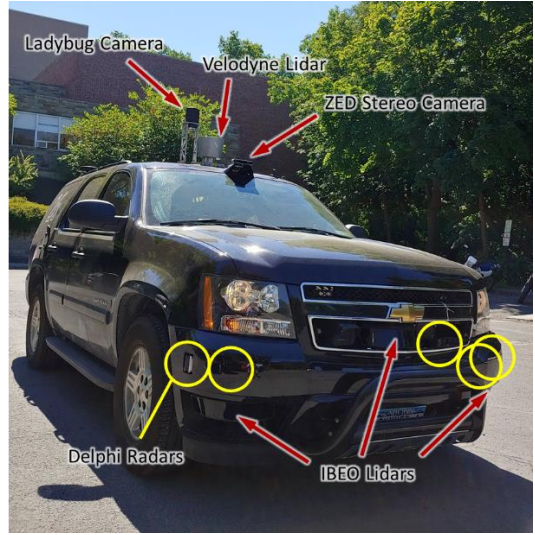

Fig. 28. Cornell University's test vehicle. All relevant cameras, radar, and lidar units are indicated.

Table 6. Descriptions of the 8 test intersections.

| Name | Description |
| --- | --- |
| Intersection A | 3-way intersection with stop sign and typically busy cross traffic |
| Intersection B | 3-way intersection with stop sign and cars typically parked along intersecting road |
| Intersection C | 4-way intersection with stop sign and cars typically parked along road past intersection |
| Intersection D | 4-way intersection with stop sign |
| Intersection E | 4-way intersection with traffic light and typically busy cross traffic |
| Intersection F | 3-way intersection with stop sign, right one-way sign, and typically busy cross traffic |
| Intersection G | 4-way intersection with stop sign and occluded views due to protruding vegetation |
| Intersection H | 4-way intersection without a stop sign or traffic light, and performed on rainy days to represent worst-case conditions |

Fig. 29 displays a summary of the experimental results for detecting the presence of an intersection. In this figure, two key metrics are plotted for each trial. The first metric is the vehicle's distance from the intersection at which the intersection is detected and its approximate relative position is estimated. An intersection is considered to be detected when its probability exceeds 0.9 and its location can be approximated. These data points are shown in red; the mean distance of the car is 34.4 m from the intersection. Intersection H does not contain any red data points because this intersection did not contain either a traffic light or stop sign, so the intersection location could not be approximated. The second metric is the vehicle's distance from the intersection at which the full intersection is detected. The full intersection is considered to be detected when the probabilities of all present intersection features (i.e. the intersection itself and all directions of travel) exceed 0.9. These data points are shown in black; the mean distance of the car is 16.2 m from the intersection. The data points in this figure are separated by intersection location, as indicated by the annotations in the figure. Thus, the 10 red data points and 10 black data points in each segment correspond to the same intersection although the test conditions differed, such as lighting, weather, traffic, and vehicle speed.

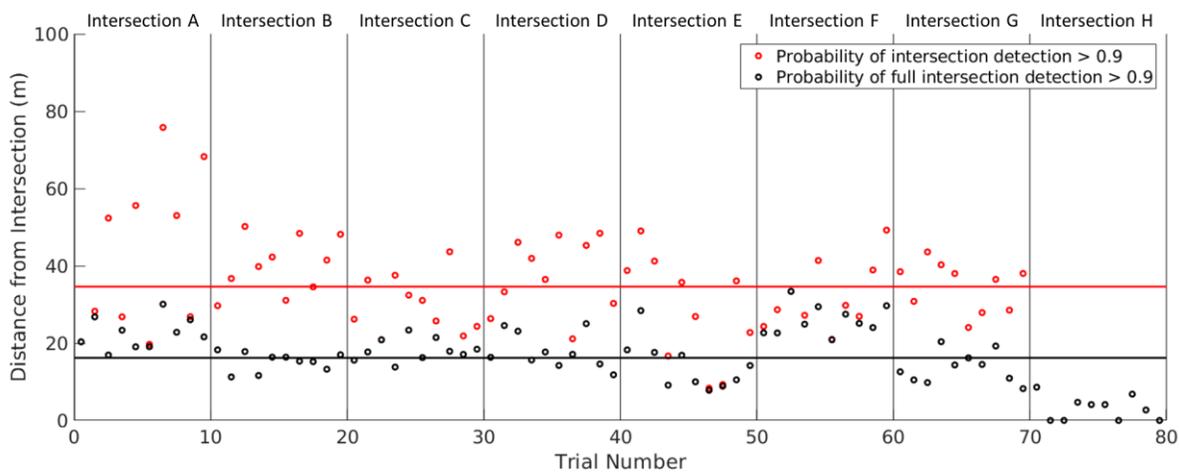

Fig. 29. Test data showing the distances from an intersection at which the test vehicle can: 1) detect the intersection and approximate its vehicle-relative location, and 2) detect the full intersection (all key intersection features needed for navigation).

Fig. 29 demonstrates that the proposed technique can successfully detect an intersection and estimate its location relative to the vehicle from approximately 34 m away, on average. Detection and localization of the intersection at this distance gives the autonomous vehicle sufficient time to prepare to turn at the intersection. Note that New York State Law states that a driver must turn on their turn signal and prepare to turn 30 meters before entering an intersection [80]. Fig. 29 also exhibits the ability to successfully detect the full intersection from approximately 16 m away, on average – a detection capability which permits the vehicle to successfully navigate the intersection.

Fig. 29 shows that the full intersection could always be successfully detected from at least 8 meters away, excluding 4 tests conducted at Intersection H. This intersection did not contain a traffic light or stop sign, and did not have any other environmental cues present as well (except for Test 1 at this intersection, which had cross traffic). Therefore, only the road surface cue could be used to detect the intersection and infer its features. Since this cue was one of the weaker cues to begin with, it is expected that solely depending on this cue would not prove to be reliable. In each of the tests at this intersection, a small number of positive detections from the road surface cue were received; however, they were not frequent and consistent enough to always produce the correct inference. To further increase the difficulty associated with this intersection, these tests were performed in the rain, and therefore, these tests represented a worst-case scenario for this algorithm. Despite the numerous difficulties associated with the tests conducted at Intersection H, the full intersection was still successfully detected in 6 of the tests, the remaining 4 tests failed to detect the full intersection. A picture showing the conditions at Intersection H is shown in Fig. 30.

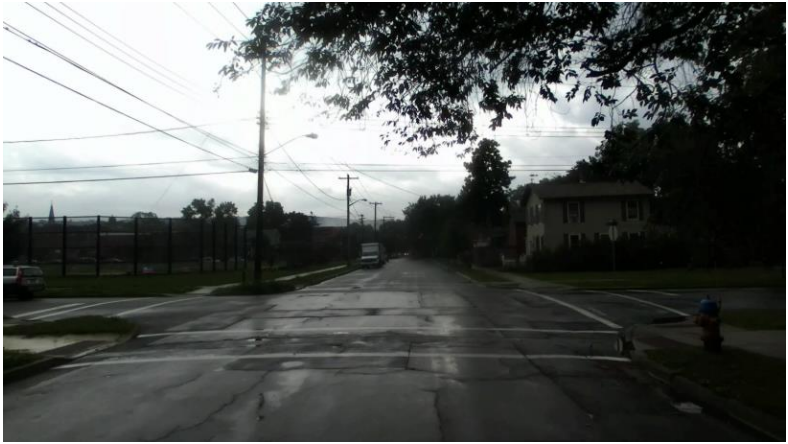

Fig. 30. Test conditions at Intersection H. This intersection did not contain a traffic light, stop sign, or cross traffic, and the tests were often conducted in the rain. This intersection represents a worst-case scenario for the proposed algorithm.

Individual tests provide further details describing the capability of this technique. Fig. 31a shows the results for test 1 at Intersection E (see Fig. 19 for reference). In this test, the vehicle was able to successfully detect the intersection at 38.6 m away, and then detect the full intersection at 19.8 m away. Note that the left turn probability increases early due to a detection of a left-facing parked car on the intersecting road. Fig. 31b shows the results for test 5 at Intersection F. The intersection was detected at 41.2 m away and the full intersection was detected at 29.4 m away. It is interesting to note that the left turn probability begins to increase at first, until a one-way sign is detected – which had been hidden behind tree branches prior to that moment – and then the probability drops sharply. The detection and tracking of cross traffic allows the right turn probability to begin to increase early in this test. Fig. 31c shows the results for test 7 at Intersection A. In this test, the intersection was detected at 75.2 m away and the full intersection was detected at 30.1 m away. This test had clear conditions, which contributed to the large detection range for this test. Moreover, the detection of cross traffic again allowed the right turn probability to increase early in the test. Finally, Fig. 31d shows the results for test 10 at Intersection B. The intersection was detected at 47.7 m away and the full intersection was detected at 17.6 m away. This test was conducted at dusk, and therefore, due to unfamiliar lighting conditions, the road segmentation did not perform as strongly as the other cases. However, due to the strong detection of left-facing parked cars along the intersecting road, the left turn probability increased sharply early in the test.

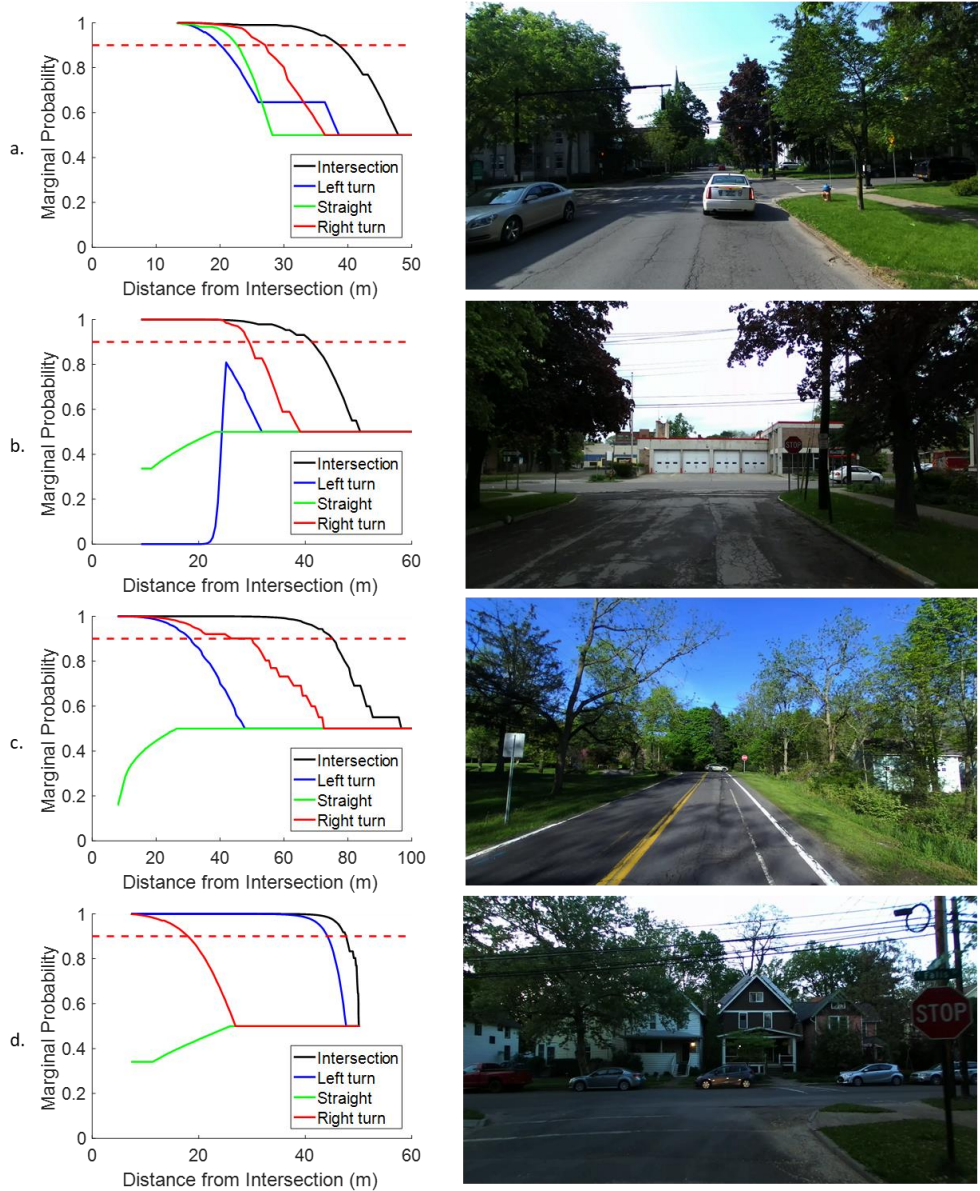

Fig. 31. Data from individual tests demonstrating strong performance of the proposed technique. The 0.9 probability threshold is indicated by the dashed red line in the left plots. (a) Test 1 at Intersection E. (b) Test 5 at Intersection F. (c) Test 7 at Intersection A. (d) Test 10 at Intersection B.

Fig. 32 shows the results from individual tests in which the proposed technique did not perform as well as the other cases. Fig. 32a shows the results for test 3 at Intersection G. The intersection in this test was detected at 43.5 m away and the full intersection was detected at only 9.8 m away. Several trees and bushes along the right side of the road obstructed the view of the right turn option, and therefore, the right turn was detected late. Fig. 32b shows the results for test 7 at Intersection E. For this test, the glare from the sun prevented the traffic light from being detected until the vehicle was nearly stopped at the intersection. Without detecting the traffic light early the intersection probability did not exceed 0.9 until late in the test. This caused the messages being sent by the cues associated with the directions of travel to remain neutral until the intersection was detected. Once the intersection was finally detected, the directions of travel could also be detected. In this test, the intersection was detected at only 8.2 m away and the full intersection was detected at 7.6 m away. Fig. 32c shows the results for test 10 at Intersection D. In this test, the intersection was detected at 30.1 m away and the full intersection was detected at 11.8 m away. This test was conducted at dusk, and therefore experienced poor performance from the road segmentation. This test also had limited cross traffic, oncoming/outgoing traffic, and cars parked along the intersecting road, which all contributed to the overall difficulty of this particular test. Overall, the vast majority of the poor performance trials seen in Fig. 29 can be attributed to one of the following unfavorable conditions: obstruction of signs and roads by vegetation, missed detections of traffic lights due to sun glare, and missed detections of road surfaces due to poor lighting conditions at night.

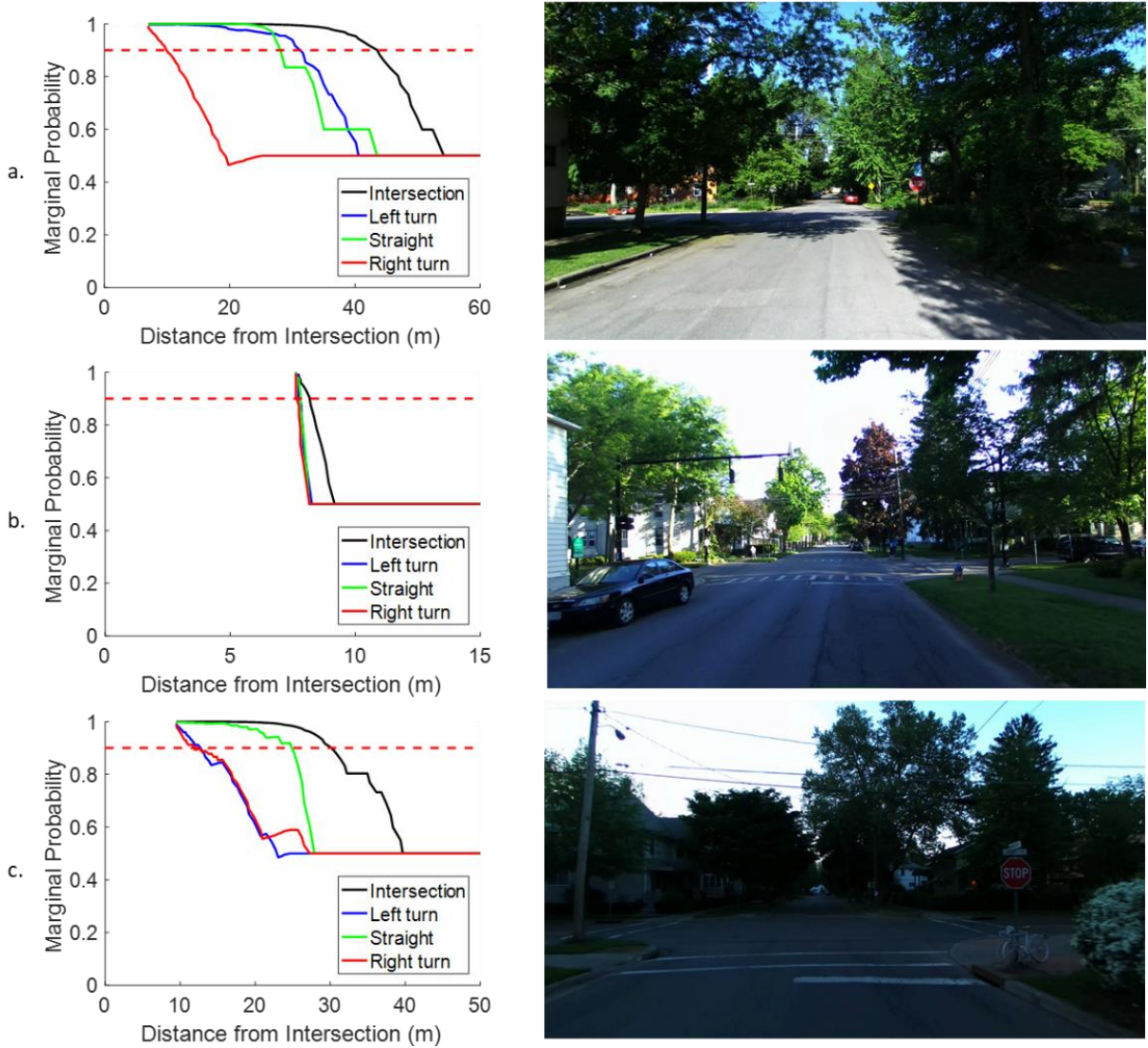

Fig. 32. Data from individual tests with weaker performance. The 0.9 probability threshold is indicated by the dashed red line in the left plots. (a) Test 3 at Intersection G. (b) Test 7 at Intersection E. (c) Test 10 at Intersection D.

## *Conclusions Drawn from Intersection Scene Estimation Study*

A novel technique is presented to address the difficulty of autonomous navigation without the use of detailed *a priori* map information. This chapter proposes a real-time static roadway scene estimation method that fuses several computer vision techniques to obtain detections of environmental cues, including traffic lights, stop signs, road surface, cross traffic, parked cars, oncoming/outgoing traffic, lane lines, one-way signs, and do not enter signs. These cues are then used to estimate the car-relative location of the intersection using belief propagation, as well as the probabilities associated with the key intersection features needed for navigation. An experimental study was conducted, which involved 80 trials at 8 different intersections under diverse conditions of lighting, weather, traffic, and vehicle speed. Results from the experimental study exhibit the capability to detect the intersection and estimate its car-relative location from an average distance of about 34 meters away, which gives an autonomous car sufficient time to make a turn. The results also show that the full intersection – with all key features needed to successfully navigate an intersection – can be detected from an average distance of approximately 16 m away. Finally, the full intersection could be successfully detected in all tests, excluding 4 worst-case scenario cases. In general, these results demonstrate that the proposed architecture can detect and locate an intersection relative to the vehicle, as well as probabilistically determine the key features of the intersection, in real-time – a capability for which other existing methods must rely on a high definition *a priori* map of the environment.

# CONCLUSIONS

In the first chapter of this dissertation, the development of the Linear Hydraulic Actuator Characterization Device (LHACD) is described. The LHACD provides a versatile and convenient experimental apparatus on which a variety of hydraulic actuators can be characterized and tested under various loading scenarios. This device was used to investigate the performance and characteristics of hydraulic artificial muscle bundles using a variable recruitment actuation and control scheme. The tests described in this study demonstrate that a variable recruitment control approach results in a significant reduction in fluidic power consumption without suffering a drop in mechanical power output, resulting in an efficiency gain of over 20%.

A novel localization and navigation architecture is presented to address the problem of autonomous driving in an urban environment in Chapter 2 of this dissertation. Without the use of prior detailed map information or GPS measurements, a pose estimation method is proposed which utilizes odometry, compass, and sparse map-based measurements to estimate the pose of the vehicle. Monte Carlo studies using a simulator developed for this work show the distance a vehicle can travel with no GPS or map information, as well as the relationship between the range of the vehicle, navigation method, landmark density, and landmark detection rate. Experimental results verify the simulation results within a small amount of deviation, as they produce a minimum range of 6.9 km for the given success rate, navigation method, landmark detection rate, and landmark density.

Chapter 3 presents a real-time static roadway scene estimation method to address the difficulty of autonomous navigation without the use of detailed *a priori* map information. This method utilizes detections of several roadway cues, such as traffic lights, stop signs, road surface, cross traffic, parked cars, oncoming/outgoing traffic, lane lines, one-way signs, and do not enter signs, to estimate the probabilities associated with the key intersection features needed for navigation. Results from an experimental study, which involved 80 trials at 8 different intersections under diverse conditions of lighting, weather, traffic, and vehicle speed, exhibit the capability to detect the intersection and estimate its car-relative location from an average distance of about 34 meters away. Furthermore, the results show that the full intersection – with all key features needed for navigation – can be successfully detected in all tests, excluding 4 worst-case scenario cases. This detection of the full intersection occurred from an average distance of about 16 meters away from the intersection.

A summary of the key contributions associated with the research conducted for this dissertation can be found in the bulleted list below.

- Designed and developed a novel hydraulic actuator characterization test platform.
- Tested the device's capabilities and showed its uniqueness compared to other devices.
- Using this device, demonstrated a variable recruitment control scheme that offers a 20% efficiency gain.
- Developed a localization and navigation architecture for autonomous driving which does not depend on prior detailed map information or GPS measurements.
- Conducted field tests that show a minimum range for a given success rate, navigation method, landmark detection rate, and landmark density, which also verify simulated results.
- Developed a real-time estimation technique that uses environmental cues to detect and locate an intersection relative to the vehicle, as well as probabilistically determine the key features of the intersection, all without the use of prior detailed map information.

- Performed an experimental study that demonstrates the ability to detect the intersection and estimate its car-relative location from an average distance of about 34 meters away, as well as the ability to detect the full intersection – with all key features needed for navigation – from an average of about 16 m away.

- Overall, developed a system architecture that enables autonomous navigation in an urban environment without the use of detailed prior map information or GPS measurements – information sources that are heavily relied upon in the autonomous driving industry.